\documentclass[10pt,journal,compsoc]{IEEEtran}
\usepackage[nocompress]{cite}
\usepackage{graphicx}
\usepackage{amsmath}
\usepackage{multirow}
\usepackage{url}
\usepackage{subcaption}
\usepackage{hyperref}

\usepackage{amsmath,amsfonts,bm}

\def\eqref#1{equation~\ref{#1}}
\def\1{\bm{1}}

\def\vn{{\bm{n}}}

\def\vv{{\bm{v}}}

\def\vx{{\bm{x}}}

\def\vz{{\bm{z}}}

\DeclareMathAlphabet{\mathsfit}{\encodingdefault}{\sfdefault}{m}{sl}
\SetMathAlphabet{\mathsfit}{bold}{\encodingdefault}{\sfdefault}{bx}{n}

\begin{document}

\title{scRAE: Deterministic Regularized Autoencoders with Flexible Priors for Clustering Single-cell Gene Expression Data}

\author{Arnab~Kumar~Mondal\textsuperscript{\textasteriskcentered}, Himanshu~Asnani\textsuperscript{\textdagger}, Parag~Singla\textsuperscript{\textasteriskcentered}, and Prathosh~AP\textsuperscript{\textasteriskcentered}
\thanks{Email id: anz188380@iitd.ac.in, himanshu.asnani@tifr.res.in,
parags@iitd.ac.in, prathoshap@iitd.ac.in}% <-this % stops a space
\thanks{\textasteriskcentered~indicated authors are affiliated with IIT Delhi. \textdagger~indicated author is affiliated with TIFR, Mumbai.}%
}

% The paper headers
\markboth{IEEE/ACM Transactions on Computational Biology and Bioinformatics}%
{Mondal \MakeLowercase{\textit{et al.}}: scRAE: Deterministic RAEs with Flexible Priors for Clustering Single-cell Gene Expression Data}

\IEEEtitleabstractindextext{%
\begin{abstract}
Clustering single-cell RNA sequence (scRNA-seq) data poses statistical and computational challenges due to their high-dimensionality and data-sparsity, also known as `dropout' events. Recently, Regularized Auto-Encoder (RAE) based deep neural network models have achieved remarkable success in learning robust low-dimensional representations. The basic idea in RAEs is to learn a non-linear mapping from the high-dimensional data space to a low-dimensional latent space and vice-versa, simultaneously imposing a distributional prior on the latent space, which brings in a regularization effect. This paper argues that RAEs suffer from the infamous problem of bias-variance trade-off in their naive formulation. While a simple AE without a latent regularization results in data over-fitting, a very strong prior leads to under-representation and thus bad clustering. To address the above issues, we propose a modified RAE framework (called the scRAE) for effective clustering of the single-cell RNA sequencing data. scRAE consists of deterministic AE with a flexibly learnable prior generator network, which is jointly trained with the AE. This facilitates scRAE to trade-off better between the bias and variance in the latent space. We demonstrate the efficacy of the proposed method through extensive experimentation on several real-world single-cell Gene expression datasets. The code for our work is available at \url{https://github.com/arnabkmondal/scRAE}.
\end{abstract}
}

% make the title area
\maketitle

\IEEEraisesectionheading{\section{Introduction}}

\IEEEPARstart{S}{ingle} cell RNA sequencing (scRNA-seq) is an emerging technology that facilitates analysis of the genome or transcriptome information from an individual cell. 
Availability of large-scale scRNA-seq datasets has opened up new avenues in cancer research \cite{dalerba2011single, navin2011tumour},  embryonic development \cite{marks2010, Tang2010} and many more.
To be able to accomplish a downstream machine learning task on the scRNA-seq data, a critical first step is to learn a `compact' representation of it or reduction of the data dimensionality. However, learning such compact representations is not straightforward due to the highly-sparse nature of scRNA-seq. This  arises from a phenomenon called `dropout', where a gene is expressed at a low or moderate level in one cell but remains undetected in another cell of the same cell type \cite{Kharchenko2014}. Dropout events occur because of the stochasticity of mRNA expression, low amounts of mRNA in individual cells and shallow sequencing depth per cell of the sequencing technologies\cite{Angerer2017} used. The excessive zero counts due to dropout cause the data to be zero-inflated and only a small fraction of the transcriptome of each cell is captured effectively. Consequently, the traditional dimensionality reduction methods fail because of excessive zero expression measurement and high variation in gene expression levels among the same type of cell. Motivated by the aforementioned challenges we make the following contributions:
\begin{enumerate}
    \item We propose a novel regularized Auto Encoder (AE)-based architecture for dimensionality reduction of scRNA-seq data.
    \item Our method introduces an additional state space in the objective of regularized AEs so that the latent prior becomes learnable. 
    \item The proposed architecture flexibly trades-off between bias (prior imposition) and variance (prior learning) and facilitates operation at different points of the bias-variance curve.
    \item We demonstrate the efficacy of the proposed method in learning a compact representation via clustering and visualization task through extensive experimentation on several scRNA-seq datasets.
\end{enumerate}

\section{Related Work}

% Since the objective of our current work is to obtain a compact lower-dimensional representation of scRNA data, we briefly review the current methods on dimensionality reduction techniques (both classical and deep-learning based methods). 

\subsection{Classical Methods on Dimensionality reduction}
Clustering through Imputation and Dimensionality Reduction (CIDR) \cite{Lin2017CIDR} is a fast algorithm that uses implicit imputation approach to reduce the effect dropouts in scRNA-seq data. CIDR computes dissimilarity matrix between the imputed gene expression profiles for every pair of single cells and perform principal coordinate analysis (PCoA). Finally, clustering is performed using first few principal coordinates. Single-cell Interpretation via Multi-kernel Learning (SIMLR) \cite{Wang2017SIMLR} performs dimensionality reduction by learning a cell-to-cell similarity matrix from the input single-cell data. To learn the similarity matrix, SIMLR learns weights for multiple kernel learning. SIMLR addresses dropout events by employing a rank constraint in the learned cell-to-cell similarity and graph diffusion. For clustering, affinity propagation (AP) can be applied to the learned similarity matrix, or k-means clustering can be used in the latent space after applying SIMLR for dimension reduction. SEURAT \cite{Satija2015SEURAT} is a sequential process involving multiple steps of normalisation, transformation, decomposition, embedding, and clustering of the scRNA-seq data. Single-cell consensus clustering (SC3) \cite{Kiselev2017SC3} continually integrates different clustering solutions through a consensus approach. SC3 first apply Euclidean, Pearson and Spearman metrics to construct distance matrices between the cells. Next, the distance matrices are transformed using either PCA \cite{Jolliffe2011} or by computing the eigen-vectors of the associated graph Laplacian, followed by kmeans clustering. Finally, SC3 deploys cluster-based similarity partitioning algorithm (CSPA) to compute the consensus matrix which is clustered using hierarchical clustering with complete agglomeration. However, SC3 is computationally heavy as it uses ensemble clustering. Single-cell Aggregated From Ensemble (SAFE) \cite{SAFE} is another consensus clustering method that takes the cluster outputs of the four algorithms: SC3 \cite{Kiselev2017SC3}, CIDR \cite{Lin2017CIDR}, Seurat \cite{Satija2015SEURAT} and t-SNE \cite{tsne} + k-means as input and ensembles using three hypergraph-based partitioning algorithms: hypergraph partitioning algorithm (HGPA), meta-cluster algorithm (MCLA) and cluster-based similarity partitioning algorithm (CSPA). SAMEClustering \cite{SAMEClustering}, Single-cell RNA-seq Aggregated clustering via Mixture model Ensemble, proposes an ensemble framework that uses SC3, CIDR, Seurat, t-SNE + k-means and SIMLR to obtain individual cluster solution. Finally it chooses a maximally diverse subset of four, according to variation in pairwise Adjusted Rand Index (ARI) and solve for an ensemble cluster solution using EM algorithms. RAFSIN \cite{rafsin} utilizes random forest graphs to cluster scRNA-seq data. scPathwayRF \cite{pathway} proposes a pathway-based random forest framework for clustering single-cell expression data. LRSEC \cite{LRSEC} assumes the scRNA-seq data exist in multiple subspaces and develop an ensemble clustering framework by using low-rank model as the basic learner to find the lowest rank representation of the data in the subspace.

\subsection{Deep-Learning based Dimensionality Reduction}
Deep learning based methods perform better as compared to the traditional methods. Autoencoders \cite{hinton2006reducing} are deep neural networks consisting of an encoder network and a decoder network. The encoder network projects high dimensional data to a low dimensional latent space and the decoder network reconstructs the original data from the compressed latent code. The encoder-decoder pair is trained to minimize reconstruction error using stochastic gradient descent. Consequently, autoencoders provide an unsupervised methodology of learning compressed representation of high dimensional data. It has been shown that minimizing a regularized reconstruction error yields an encoder-decoder pair that locally characterizes the shape of the data-generating density \cite{rae_data_dist}. Single-cell Variational Inference (scVI) \cite{Lopez2018SCVI} is an autoencoder based fully probabilistic approach to normalize and analyze scRNA-seq data. It adapts hierarchical Bayesian model and parameterizes conditional distributions using deep neural networks. Sparse Autoencoder for Unsupervised Clustering,
Imputation, and Embedding (SAUCIE) \cite{amodio2019exploring} is a sparse autoencoder for unsupervised clustering, imputation and embedding. DR-A \cite{Lin2020} implements a deep adversarial variational autoencoder based framework. They propose a novel architecture Adversarial Variational AutoEncoder with Dual Matching (AVAE-DM). An encoder-decoder pair learns to project scRNA-seq data to a low dimensional manifold and reconstructs. Two discriminator are trained in the data-space and latent-space respectively to discriminate between real and fake samples. scVAE \cite{scVAE} adapts the framework of Variational AutoEncoder (VAE) \cite{kingma2013autoencoding} for analysing scRNA-seq data. scVAE makes use of likelihood function based on zero inflated Poisson distribution and zero inflated negative binomial distribution to model `dropout' event in scRNA-seq data. scVAE assumes either a Gaussian or a mixture of Gaussian prior.
scDeepCluster \cite{Tian2019scDeepCluster} combines Deep Embedding clustering \cite{dec, improvedDEC} with denoising Deep Count Autoencoder (DCA) \cite{Eraslan2019DCA} to analyze and cluster scRNA-seq data. scziDesk \cite{scziDesk} extends the idea of scDeepCluster \cite{Tian2019scDeepCluster} by introducing weighted soft K-means clustering with inflation operation. GraphSCC \cite{zeng2020accurately} exploits graph   convolutional network \cite{GCN} to cluster cells based on scRNA-seq data by accounting structural relations between cells. scGNN\cite{scGNN} is another framework that exploits graph neural networks to formulate and aggregate inter-cell relationships and provide a hypothesis-free deep learning framework for scRNA-Seq analyses. Contrastive-sc \cite{ciortan2021contrastive} is a self-supervised algorithm for clustering scRNA sequence. 
% scSemiCluster \cite{scSemiCluster} combines the reference data and target data for training and leverage structural similarity regularization on the reference domain to restrict the clustering solutions of the target domain. 
scDCC \cite{tian2021model} incorporates domain knowledge as prior information into the modeling process to guide the deep neural framework to simultaneously learn more informative latent representations and biologically meaningful clusters.
\subsection{Bias-Variance Trade-off in RAEs}
Deep-learning based representation learning techniques discovers useful low-dimensional features of high-dimensional data. Auto-encoder based generative models such as VAE \cite{kingma2013autoencoding} and its variants \cite{Lopez2018SCVI, scVAE} or AAE \cite{AAE} and its variants \cite{Lin2020, MaskAAE, FlexAE} are a few examples of deep learning based models that are used for dimensionality reduction. They do so by learning a projection from high-dimensional data-space to a low-dimensional latent space (encoder) and an inverse projection from low dimension representation space to the original data-space (decoder). The encoded latent space is also regularized to conform to some known primitive distribution. However, this regularization increases bias in the network and make learning good representation difficult. On the other hand, if the latent space is completely unregulated the model might memorize unique codes per sample in the finite data regime. In other words, the learnt latent space becomes non-smooth, resulting in increased variance (over-fitting) of the model. This is the infamous bias-variance trade-off that warrants a flexible prior which could facilitate the operation of an AE-model at different points of the bias-variance curve which forms the basis for our proposed work. 

\section{Proposed Method}
% We adopt AE-based generative approach to directly model gene expression from scRNA-seq data. To begin with, we mathematically define different components of a RAE and briefly review latent variable generative models in this section. Next we discuss our proposed method and its implementation.
\subsection{Regularized Generative Autoencoders}
An AE based generative model combines the task of inference and generation by modelling the joint distribution over high-dimensional data space, $\mathcal{X}$ and low-dimensional latent space, $\mathcal{Z}$. The joint inference distribution is defined as $Q_\phi(\vx, \vz)=P_d(\vx)Q_\phi(\vz|\vx)$ and the joint generative distribution is defined as $P_\theta(\vx, \vz)=Q_\psi(\vz)P_\theta(\vx|\vz)$,
where, $P_d(\vx)$ represents the true data distribution over $\mathcal{X}$. $Q_\phi(\vz\lvert\vx)$ is the posterior distribution of the latent code given data point. $Q_\phi(\vz)=\int P_d(\vx)Q_\phi(\vz\lvert\vx)d\vx$, is the aggregated posterior. $Q_\phi(\vz\lvert\vx)$ is parameterized by a neural network called Encoder, $E_\phi$, that projects the input data $\vx \in \mathcal{X}$ to a latent code $\vz \in \mathcal{Z}$. $p_\theta(\vx|\vz)$ is parameterized by another neural network called Decoder, $D_\theta$, which learns inverse mapping from $\vz \in \mathcal{Z}$ to $\vx \in \mathcal{X}$. $P_Z(\vz)$ denotes the prior distribution over the latent space, $\mathcal{Z}$. $\phi \in \Phi$, and $\theta \in \Theta$ are vectors of learnable parameters. Mathematically, a RAE solves the following optimization objective:
\begin{equation}
\begin{gathered}
\mathop{\inf}_{\phi, \theta}\Bigg(\mathop{\mathbb{E}}_{P_d(\vx)}\mathop{\mathbb{E}}_{Q_\phi(\vz|\vx)}\bigg[c\Big(\vx, D_\theta\big(E_\phi(\vx)\big)\Big)\bigg]\Bigg)
\\\text{such that } Q_\phi(\vz) = P_Z(\vz)
\label{eqn:rae_constrained_obj}
\end{gathered}
\end{equation}
Where, $c:\mathcal{X} \times \mathcal{X} \to \mathbb{R}^+$ denotes any measurable cost function (such as Mean Square Error (MSE) and Mean Absolute Error(MAE)). The remaining notations have their usual meaning as defined above.
The above constrained optimization objective can equivalently be written as an unconstrained optimization problem by introducing a Lagrangian:
\begin{gather}
\begin{split}
D_{RAE} &= \mathop{\inf}_{\phi, \theta}\Bigg(\underbrace{\mathop{\mathbb{E}}_{P_d(\vx)}\mathop{\mathbb{E}}_{Q_\phi(\vz|\vx)}\Big[c\Big(\vx, D_\theta\big(E_\phi(\vx)\big)\Big)\Big]}_{\text{a}} +\\ &\qquad\lambda \cdot \underbrace{\vphantom{\mathop{\mathbb{E}}_{Q_\phi(\vz|\vx)}}D_{Z}\big(Q_\phi(\vz),P_Z(\vz)\big) }_\text{b}\Bigg)
\end{split}
\label{eqn:rae_relaxed_obj}
\end{gather}
Where, $D_{Z}(.)$ denotes any divergence measure such as Kullaback-Leibler, Jenson-Shannon or Wasserstein distance, between two distributions, $\lambda$ is the Lagrange multiplier, and rest of the symbols have their usual meaning as defined before.

\subsection{scRAE}

\begin{figure*}[!ht]
    \centering
    \includegraphics[clip, trim=50 230 80 40, keepaspectratio,width=\textwidth]{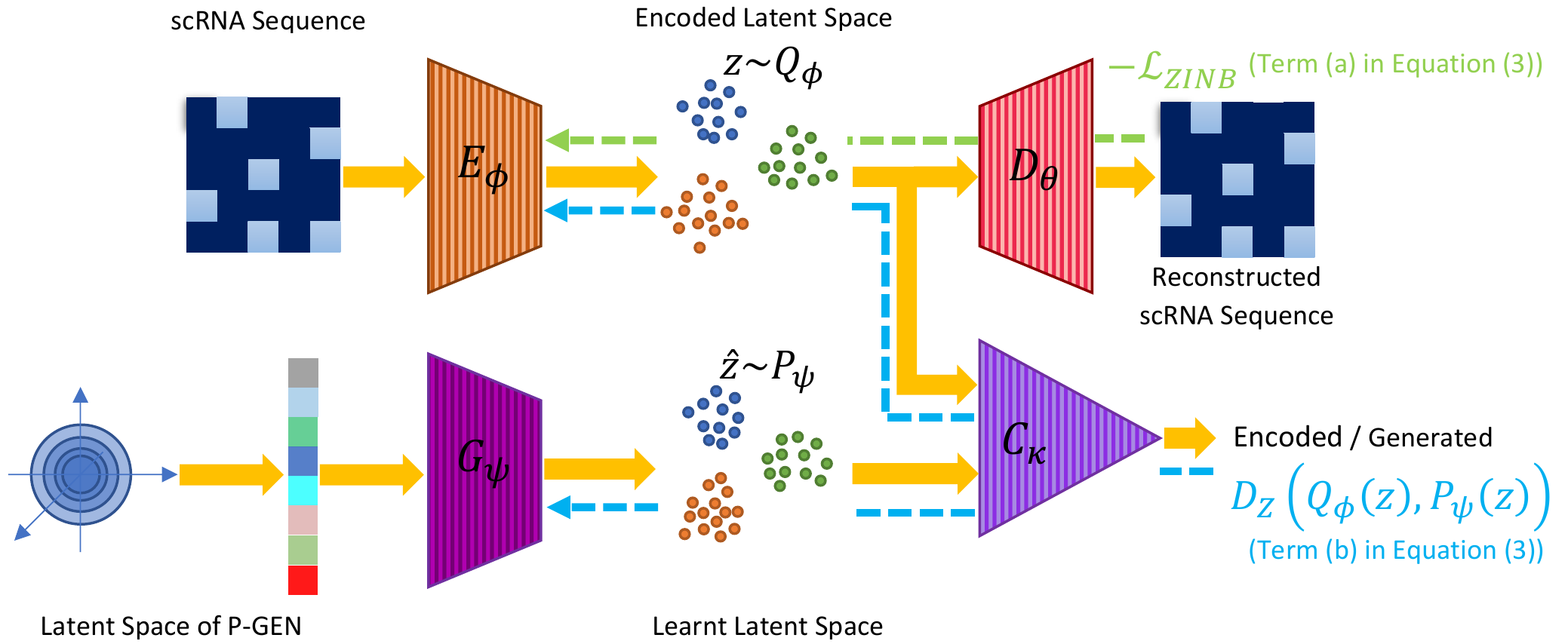}
    \caption{The novel architecture of scRAE consists of a reconstruction pipeline (the autoencoder) and a P-GEN network. The high-dimensional sparse single cell RNA sequence is mapped to a dense low dimensional latent representation using the encoder network, $E_\phi$ and the original sequence is reconstructed using a decoder network, $D_\theta$. The encoded latent space is constrained by the P-GEN network which consists of a latent generator network, $G_\psi$ and a critic, $C_\kappa$. $G_\psi$ learns the prior flexibly based on feedback from $C_\kappa$. The learnt latent representations are clustered corresponding to different cell types in the dataset. The entire network is trained in an end-to-end fashion. Solid yellow arrows in the left-right direction illustrates the forward path and the color coded dashed arrows in the right-left direction illustrates the backward flow of gradients due to different terms in the optimization objective.}
    \label{fig:bd}
    \vspace{-2mm}
\end{figure*}

In this work, we argue that when the prior is extremely simple such as Gaussian, it increases the bias in the network and prevent the autoencoder from discovering the true structure. On the other hand, when the latent space is not restricted, due to increased variance the network tend to memorize the training samples leading to overfitting. As a remedial measure we propose to learn the prior jointly and flexibly along with AE training.\par
As described in Figure \ref{fig:bd}, scRAE consists of four neural nets. The Encoder network, $E_\phi$ maps high dimensional sparse scRNA sequence to a low dimensional dense latent code, $\vz\sim Q_\phi$ and the Decoder network, $D_\theta$ learns an inverse mapping. These two networks together constitute the reconstruction pipeline and are trained to minimize a reconstruction loss. The Generator network, $G_\psi$ and the critic network, $C_\kappa$ forms the P-GEN network, which simultaneously learns to bring $P_\psi$ closer to $Q_\phi$ and regularizes the encoded latent space.\par
scRAE is trained in an end-to-end fashion. The encoded aggregated posterior distribution, $Q_\phi(\vz) = \int Q_\phi (\vz|\vx) P_d(\vx) d\vx$ acts as the target prior for the generator network, $G_\psi$. On the other hand, the learnt prior distribution, $P_\psi(\vz)$ regularizes the learnt latent space. The objective function of scRAE is:
\begin{equation}
\begin{split}
D_{scRAE}&=\mathop{\inf}_{\psi, \phi, \theta}\Bigg(\underbrace{\mathop{\mathbb{E}}_{P_d(\vx)}\mathop{\mathbb{E}}_{Q(\vz|\vx)}\Big[c(\vx, D_\theta(\vz)\Big]}_{\text{a}} + \\
&\qquad\lambda \cdot \underbrace{ \vphantom{\mathop{\mathbb{E}}_{P(\vx)}} D_{Z}(Q_\phi(\vz)||P_\psi(\vz)) }_\text{b}\Bigg)
\end{split}
\label{eqn:scRAE_obj}
\end{equation}, where $\lambda$ denotes the Lagrange multiplier\footnote{Theoretically, the objective should be optimized w.r.t. the Lagrange multiplier $\lambda$. However, in practical implementations \cite{WAE} it is considered to be a hyper-parameter.}, and the remaining notations are as defined before. \par
One way to visualize Equation \ref{eqn:scRAE_obj} is that the objective is to minimize a reconstruction error (term a) regularized by a distributional divergence (term b).
Note that, AAE \cite{AAE} and WAE \cite{WAE} are special cases of scRAE when the generator network, $G_\psi$ is an identity function.\par

\subsection{Realization of scRAE}

For implementation, we use Zero-inflated negative binomial (ZINB) based negative log-likelihood objective for $c$ in term (a) of Eq. \ref{eqn:scRAE_obj}. $D_Z$, in principle can be chosen to be any distributional divergence such as Kullback-Leibler divergence (KLD), Jensen–Shannon divergence (JSD), Wasserstein Distance and so on. In this work, we propose to use Wasserstein distance and utilize the principle laid in \cite{arjovsky2017wasserstein, gulrajani2017improved}, to optimize the divergence (term (b) in Equation \ref{eqn:scRAE_obj}). The loss functions used for different blocks of FlexAE are as follows:
\begin{enumerate}
    \item {Likelihood Loss - Realization of Term a in Eq. \ref{eqn:scRAE_obj}: Zero-inflated negative binomial (ZINB) model is suitable for modeling count variables with excessive zeros and it is usually used for overdispersed count variables. In order to handle `dropout' event in scRNA-seq data, $P_\theta(\vx|\vz)$ is modelled as zero inflated negative binomial (ZINB) distribution in scRAE. ZINB is defined as:
    \begin{equation}
        f_{ZINB}(x) = \begin{cases}
        \pi + (1 - \pi)f_{NB}(x) ~if~x=0\\
        (1 - \pi)f_{NB}(x) ~ if~ x > 0
        \end{cases}
    \end{equation}
    \begin{equation}
        f_{NB}(x) = \frac{\Gamma (x + \alpha^{-1})}{\Gamma(\alpha^{-1})\Gamma(x+1)}\bigg(\frac{1}{1+\alpha\mu}\bigg)^{\frac{1}{\alpha}}\bigg(\frac{\alpha\mu}{1+\alpha\mu}\bigg)^{x}
    \end{equation}, where $\pi$ denotes the probability of excessive zero, $\alpha$ is the dispersion parameter, $\mu$ denotes mean. If, logit, $l = \log \frac{\pi}{1 - \pi}$, the log-likelihood of ZINB distribution is defined as follows:
    \begin{equation}
        \mathcal{L}_{ZINB} = \begin{cases}
        \sum -Softplus(-l) + Softplus\Bigg(\\\qquad-l + \frac{1}{\alpha}\log\bigg(\frac{1}{1+\alpha\mu}\bigg)\Bigg)~if~x=0 \\
        \sum -Softplus(-l)-l+\\\qquad\log\big(f_{NB}(x)\big) ~~~~~~~~~~~if~ x > 0
        \end{cases}
    \end{equation}
    The encoder-decoder pair is trained to minimize the negative log-likelihood, $-\mathcal{L}_{ZINB}$ defined under ZINB distribution as reconstruction loss.
    \begin{equation}
    L_{Reconstruction} = -\mathcal{L}_{ZINB}
    \label{eqn:nll}
    \end{equation}}
    From implementation point of view, $\mu$, $\frac{1}{\alpha}$ and $l$ are learnable parameters. Specifically, if $h_\theta$ denotes the final hidden layer of the decoder network, two fully connected dense layers are used to learn the logit, $l$, and mean, $\mu$. The dispersion parameter, $\frac{1}{\alpha}$ is learnt as a standalone vector.
    \begin{equation}
        \mu = \exp(s) \times Softmax(W_\mu h_\theta)
    \end{equation}, where $s \sim \mathcal{N}(\mu_s, \sigma_s)$. $\mu_s$ and $\sigma_s$ are one-dimensional outputs of the encoder network. The network is trained to match $\mathcal(\mu_s$, $\sigma_s)$ to $\mathcal{N}(\mu_G, \sigma_G)$ by minimizing the KL divergence between two Gaussian distributions. $\mu_G$ and $\sigma^2_G$ are respectively the mean and the variance of the log library size. Note, that the choice of activation ensures the mean, $\mu$ is always non-negative.
    \begin{equation}
        \alpha^{-1} = \exp(\vv)
    \end{equation}, where $\vv$ is a randomly initialized independent learnable vector. Like $\mu$, the dispersion parameter is also non-negative by design choice.\\
    Finally, the logit, $l$ is the final output of decoder
    \begin{equation}
        l = W_l h_\theta
    \end{equation}
    The loss function for the autoencoder can now be written as:
    \begin{equation}
        L_{AE} = \mathcal{L}_{ZINB} + \lambda D_{KL}\big(\mathcal{N}(\mu_s, \sigma_s)||\mathcal{N}(\mu_G, \sigma_G)\big)
        \label{eqn:ae_loss}
    \end{equation}

  \item {Wasserstein Loss - We use Wasserstein distance \cite{arjovsky2017wasserstein} for $D_Z$ (Term b Eq. \ref{eqn:scRAE_obj}):}
    \begin{equation}
        \begin{split}
            L_{Critic} &= \frac{1}{s}\sum_{i=1}^{s}C_\kappa(\hat{\vz}^{(i)}) - \frac{1}{s}\sum_{i=1}^{s}C_\kappa (\vz^{(i)}) +\\&\qquad \frac{\beta}{s}\sum_{i=1}^{s}\big(\lvert\lvert\nabla_{\vz_{avg}}^{(i)}C_\kappa (\vz_{avg}^{(i)})\lvert\lvert - 1\big)^2
        \end{split}
        \label{eqn:critic_loss}
    \end{equation}
    \begin{equation}
            L_{Gen} = -\frac{1}{s}\sum_{i=1}^{s}C_\kappa(\hat{\vz}^{(i)})
            \label{eqn:gen_loss}
    \end{equation}
    \begin{equation}
            L_{Enc} = \frac{1}{s}\sum_{i=1}^{s}C_\kappa(\vz^{(i)})
            \label{eqn:enc_loss}
    \end{equation}
  
\end{enumerate}
Where, $\vz^{(i)} = E_{\phi}(\vx^{(i)})$, $\hat{\vz}^{(i)} = G_{\psi}(\vn^{(i)})$ and $\vn^{(i)} \sim \mathcal{N}(0, I)$. $\vz_{avg}^{(i)} = \alpha\vz^{(i)} + (1-\alpha)\hat{\vz}^{(i)}$,  $\alpha,\beta$ are hyper parameters, with $\alpha \sim \mathcal{U}[0, 1]$, and $\beta$ as in  \cite{gulrajani2017improved}. $E_{\phi}, D_\theta, G_\psi$, and $C_\kappa$ denote the encoder, decoder, latent generator and critic respectively.\par
As mentioned above, the auto-encoder is required to be optimized jointly with the P-GEN to ensure regularization in the AE latent space. This regularization effectively enforces smoothness in the learnt latent space and prevents the AE from  overfitting on the training examples. In order to be able to satisfy the above requirement in practice, we optimize each of the four losses (Equation \ref{eqn:ae_loss} - \ref{eqn:enc_loss}) specified above in every training iteration. Specifically, in each learning loop, we optimize the $L_{AE}$, $L_{Gen}$, $L_{Enc}$, and $L_{Critic}$ in that order using a learning schedule. We use Adam optimizer for our optimization. To stabilize the adversarial training, we utilize the principle outlined in WGAN-GP\cite{gulrajani2017improved}.

% \begin{figure*}[!ht]
%     \centering
%     \includegraphics[trim=40 390 20 35, clip, keepaspectratio,width=\textwidth]{assets/preprocessing.pdf}
%     \caption{Flowchart view of the three steps involved in scRNA sequence preprocessing as outlined in \cite{amezquita2020orchestrating}.}
%     \label{fig:preprocessing}
% \end{figure*}

\section{Experiments and Results}
\begin{table}[!t]
    \caption{Summarized description of datasets used.}
    \begin{center}
    \resizebox{\columnwidth}{!}{
    \begin{tabular}{c c c c c c c}
         \hline
         Dataset & Protocol & Sample Size & \# of Genes & \# Cell Types\\
         \hline\hline
         Klein-$2k$ \cite{klein2015droplet} & inDrop & $2,717$ & $24,175$ & $4$ \\
         Han-$2k$ \cite{HAN2018MicorwellSeq} & Microwell-seq & $2,746$ & $19,079$ & $16$ \\
         Zeisel-$3k$ \cite{Zeisel1138} & STRT-Seq UMI & $3,005$ & $19,972$ & $9$ \\
         Segerstolpe-$3k$ \cite{segerstolpe2016single} & Smart-Seq2 & $3,514$ & $25,525$ & $13$ \\
         Cao-$4k$ \cite{Cao2017} & sci-RNA-seq & $4,186$ & $11,955$ & $10$ \\
         Baron-$8k$ \cite{baron2016single} & inDrop & $8,569$ & $20,125$ & $14$ \\
         Macosko-$44k$ \cite{Macosko20151202} & Drop-seq & $44,808$ & $23,288$ & $39$\\
         Zheng-$68k$ \cite{Zheng2017} & 10X & $68,579$ & $32,738$ & $10$\\
         Zheng-$73k$ \cite{Zheng2017} & 10X & $73,233$ & $32,738$ & $8$\\
         Rosenberg-$156k$ \cite{Rosenberg176} & SPLiT-Seq & $156,049$ & $26,894$ & $73$\\
         \hline
    \end{tabular}
    }
    \end{center}
    \label{tab:dataset_summary}
\end{table}    
To evaluate the efficacy of the proposed method, we have conducted two types of experiments using the lower dimensional embedding of high dimensional scRNA data: 1. Clustering (Refer section \ref{sec:clustering_expt}) and 2. Visualization (Refer section \ref{sec:data_vis_expt}). In both the experiments, the training is completely unsupervised as no label information is used. However, for performance evaluation and visualization the test labels are used. Further to illustrate the scalability of the proposed method, we use ten real datasets of varying sample size as described in the section \ref{sec:datasets}.

\subsection{Datasets}\label{sec:datasets}
In this paper, the following ten datasets are considered for the evaluation.
\begin{enumerate}
    \item \textbf{Klein-$2k$} \cite{klein2015droplet}: This dataset has $2,717$ single-cell transcriptomes with $24,175$ features from mouse embryonic stem cells. The cell labels for this dataset can only be considered `silver standard' as labels were assigned using computational methods and the authors' knowledge of the underlying biology.
    \item \textbf{Han-$2k$} \cite{HAN2018MicorwellSeq}: The authors in \cite{HAN2018MicorwellSeq} constructed a `mouse cell atlas' with more than  $400,000$ single cells covering all of the major mouse organs. In this paper, we consider $2,746$ samples from mouse bladder cells having $16$ distinct classes.
    \item \textbf{Zeisel-$3k$} \cite{Zeisel1138}: This dataset consists of $3,005$ single cell transcriptomes from the primary somatosensory cortex (S1) and thehippocampal CA1 region of the juvenile mouse brain. As reported in \cite{Zeisel1138}, the dataset has nine major classes and forty seven distinct sub-classes comprising all known major cell types in the cortex region.
    \item \textbf{Segerstolpe-$3k$} \cite{segerstolpe2016single}: The authors of \cite{segerstolpe2016single} sequenced the transcriptomes of human pancreatic islet cells from healthy and type 2 diabetic donors. The dataset consists of $3,514$ samples with $25,525$ features.
    \item \textbf{Cao-$4k$} \cite{Cao2017}: This dataset has of $4,186$ samples with $11,955$ features or genes from worm neuron cells. The expressions are categorized into $10$ distinct classes.
    \item \textbf{Baron-$8k$} \cite{baron2016single}: This dataset has $8,569$ samples consisting of $20,125$ genes from human pancreatic islet. Cells can be divided into 14 clusters viz. `acinar', `activated stellate', `alpha', `beta', `delta', `ductal',`endothelial', `epsilon', `gamma', `macrophage', `mast', `quiescent stellate', `Schwann', and `t-cell'.
    \item \textbf{Macosko-$44k$} \cite{Macosko20151202}: This dataset contains single-cell gene expression counts from $39$ types of $44,808$ mouse retinal cells.
    \item \textbf{Zheng-$68k$} \cite{Zheng2017}: This dataset is comprised of $68,579$ single-cell transcriptomes of fresh peripheral blood mono-nuclear cells (PBMC) in a healthy human. The dataset has $10$ different types of cells.
    \item \textbf{Zheng-$73k$} \cite{Zheng2017}: This dataset is created by combining $8$ separate datasets of different purified cell types. In this dataset, there are $73,233$ sequences from $8$ distinct cell types.
    \item \textbf{Rosenberg-$156k$} \cite{Rosenberg176}: This dataset contains $156,049$ single-cell transcriptomes from postnatal day $2$ and $11$ mouse brains and spinal cords. $73$ distinct cell types were identified and reported in \cite{Rosenberg176}.
\end{enumerate}
Each dataset has been preprocessed and randomly partitioned into $80\%$ training and $20\%$ test samples. The following section describes the preprocessing pipeline.\par
\subsection{Preprocessing and Gene Selection}\label{sec:preprocessing}
Data preprocessing is the first crucial step in preparing the raw count data to suit the proposed framework scRAE. As outlined in \cite{amezquita2020orchestrating}, preprocessing of scRNA sequence includes the following three steps:

\begin{table*}[!ht]
\begin{center}
\caption{Comparison of the average normalized mutual information across $10$ real datasets achieved by different algorithms.}
\label{tab:average_nmi}
\begin{tabular}{cccccccc}
\hline
$n_z$ & scRAE           & scziDesk \cite{scziDesk} & scVAE \cite{scVAE}  & scDeepCluster \cite{Tian2019scDeepCluster} & DR-A \cite{Lin2020}   & SAUCIE \cite{amodio2019exploring} & scVI \cite{Lopez2018SCVI}  \\
\hline\hline
2    & \textbf{0.6975} & 0.6259   & 0.5755 & 0.5808        & 0.6441 & 0.4450 & 0.6164 \\
10   & \textbf{0.7093} & 0.6427   & 0.5887 & 0.6139        & 0.6744 & 0.4826 & 0.6271 \\
20   & \textbf{0.7126} & 0.6489   & 0.6132 & 0.6277        & 0.6823 & 0.4787 & 0.6405\\
\hline
\end{tabular}
\end{center}
\vspace{-3mm}
\end{table*}

\begin{enumerate}
    \item \textbf{Selection and filtration of cells and genes for quality control:} % Sometimes, cells might get damaged during processing, and the sequencing protocol might capture such cells partially. These low-quality cells would intervene in downstream analyses. Similarly, genes expressed in less than a few cells might be considered as noise. 
    We eliminate each cell without at least one expressed gene and each gene expressed in less than ten cells to reduce the effect of low-quality cells and noisy expression in downstream analyses.
    \item \textbf{Data normalization and scaling:} % Normalization and scaling eliminate cell-specific biases and allow us to perform explicit comparisons across cells. We normalize each cell/sample by total counts over all genes, such that after normalization every cell has the same total count. 
    We exclude very highly expressed genes from the computation of the normalization factor for individual cell; because these highly expressed genes would strongly influence the resulting normalized values for all other genes \cite{SPRING}. Further, the normalized count data is log-transformed to adjust for the mean-variance relationship.
    \item \textbf{Feature/Gene selection:} Finally, a subset of high-variance attributes is selected for downstream analysis by modeling the variance across the cells for every gene and retaining the highly variable genes. This feature selection step reduces computational burden and noise from uninformative genes. Following \cite{Lin2020}, We select $720$ genes (features) that exhibit the highest inter-cell variance. Refer to Section \ref{sec:feature_dim_vs_nmi} for an ablation study to understand the impact of feature dimension. We have used the dispersion-based method outlined in \cite{Satija2015SEURAT} termed as Highly Variable Gene (HVG) selection method. However, we have also experimented with two other popular methods such as SCMarker \cite{SCMarker}, and M3Drop \cite{M3Drop} as detailed in Section \ref{gene_sel}.
\end{enumerate} 

\begin{figure*}[t!]
    \centering
    \begin{subfigure}[t]{\textwidth}
        \centering
        \includegraphics[trim={2 4 2 2}, clip, keepaspectratio, width=\textwidth]{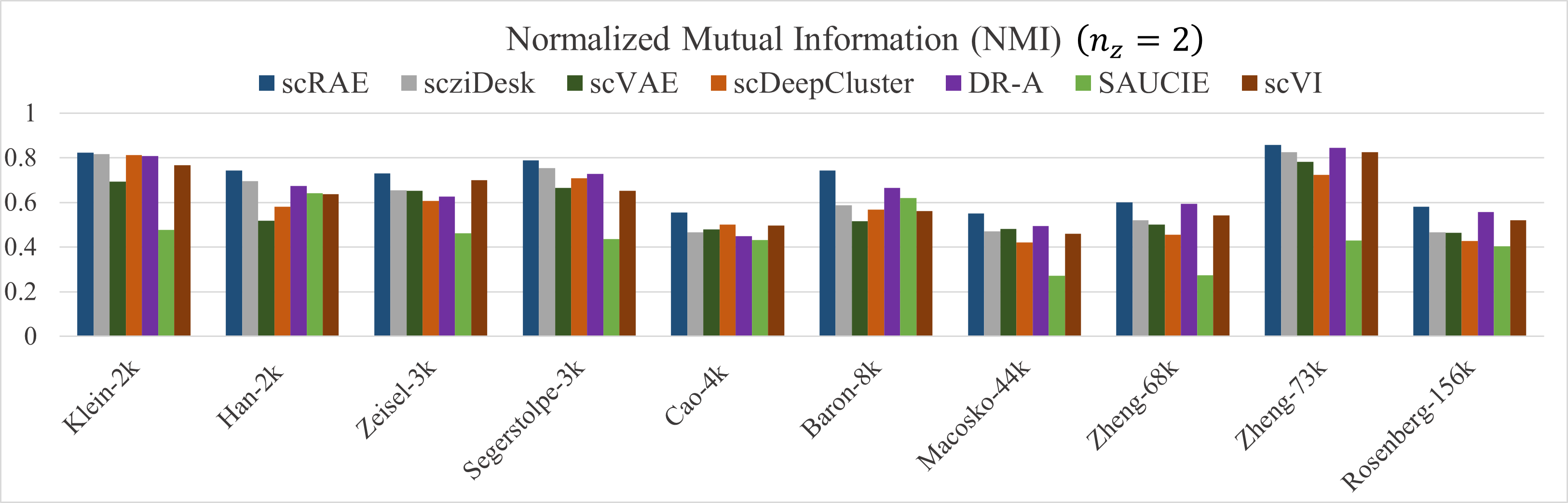}
        \caption{}
        \label{fig:nmi_nz_2}
    \end{subfigure}%
    \hfill
    \begin{subfigure}[t]{\textwidth}
        \centering
        \includegraphics[trim={2 4 2 2}, clip, keepaspectratio, width=\textwidth]{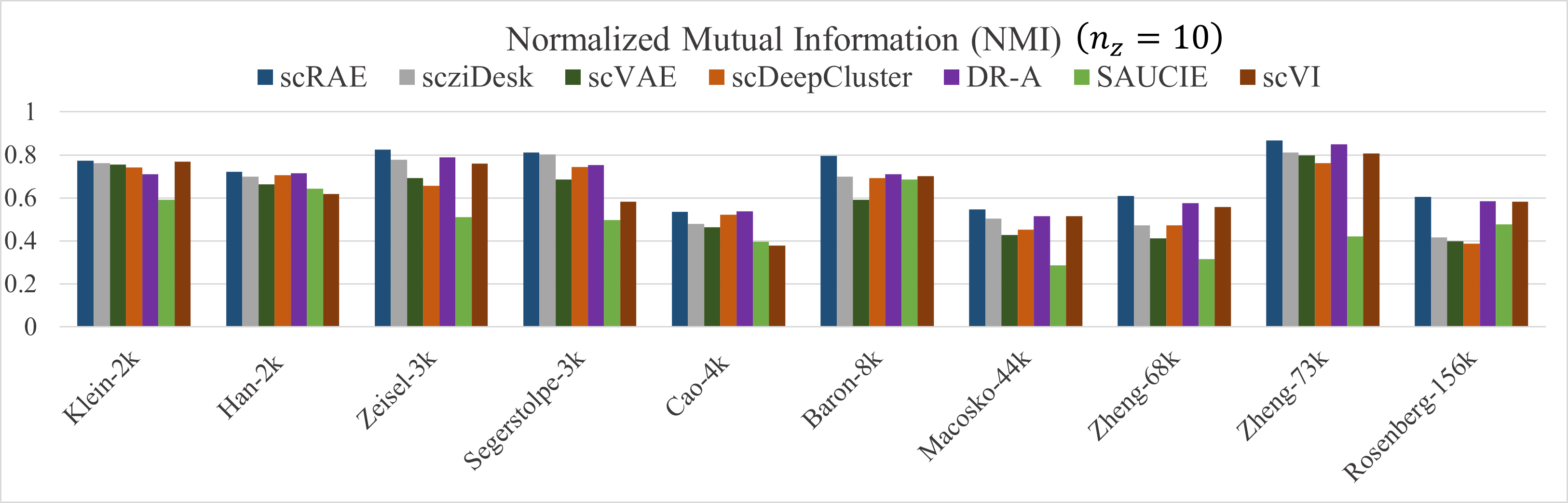}
        \caption{}
        \label{fig:nmi_nz_10}
    \end{subfigure}%
    \hfill
    \begin{subfigure}[t]{\textwidth}
        \centering
        \includegraphics[trim={2 4 2 2}, clip, keepaspectratio, width=\textwidth]{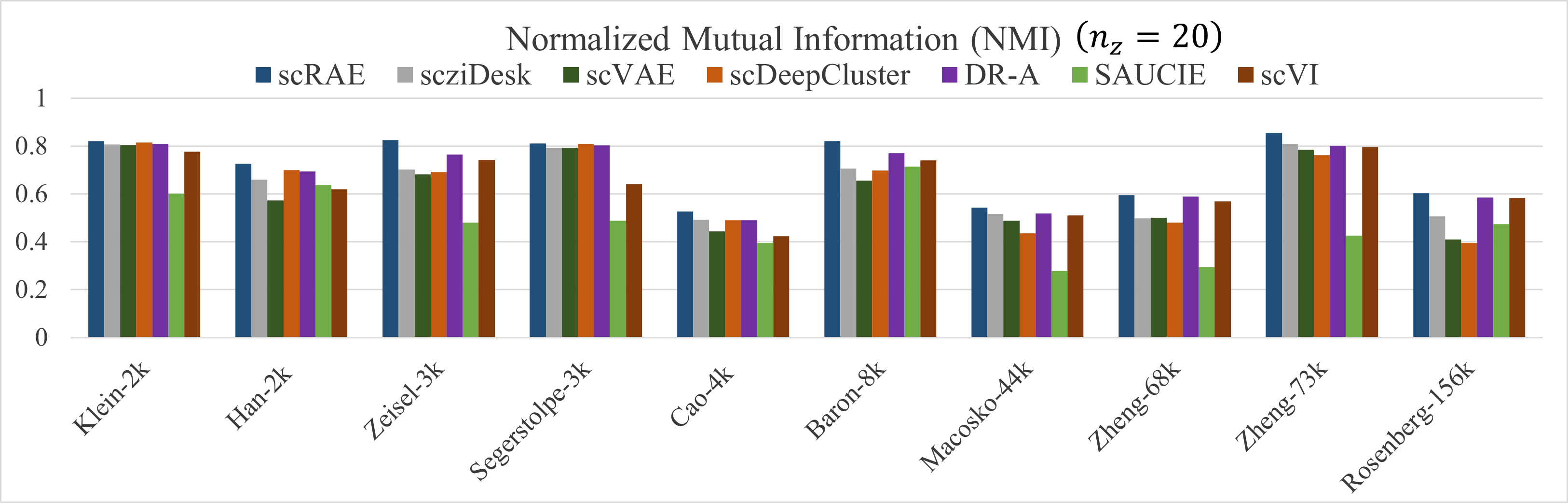}
        \caption{}
        \label{fig:nmi_nz_20}
    \end{subfigure}%
    \caption{Clustering performance of scRAE is compared against $6$ baseline methods. As can be seen from the figure, scRAE achieves highest NMI score for all of the ten datasets irrespective of the latent dimension. Higher NMI score indicates better cluster purity and better performance w.r.t. ground truth labels.}
    \label{fig:nmi_comparison}
    % \vspace{-3mm}
\end{figure*}

\subsection{Clustering}\label{sec:clustering_expt}
In order to benchmark the proposed algorithm scRAE\footnote{\url{https://github.com/arnabkmondal/scRAE}}, we compare its performance against $6$ state-of-the-art deep learning based methods as listed below:
scVI \cite{Lopez2018SCVI}, SAUCIE \cite{amodio2019exploring}, scDeepCluster \cite{Tian2019scDeepCluster}, scVAE \cite{scVAE}, DR-A \cite{Lin2020}, scziDesk \cite{scziDesk}.
We conduct experiments using ten real-world datasets as described in Section \ref{sec:datasets}. To assess the effectiveness of the proposed method, we evaluate the impact of the learnt representations on the performance of K-means clustering algorithm. First, the gene expression data are compressed using the proposed method and the baseline methods. Next, K-means clustering algorithm is used to compute cluster assignment on the compressed representation. To quantitatively evaluate the quality of clustering, we compute and report the normalized mutual information (NMI) scores \cite{NMI} in Figure \ref{fig:nmi_comparison} as in previous works such as DRA\cite{Lin2020}, scDeepCluster\cite{Tian2019scDeepCluster}. NMI is defined as follows:
\begin{equation}
    \text{NMI} = \frac{I(Y_{true}; Y_{pred})}{\sqrt{H(Y_{true}) H(Y_{pred})}}
\end{equation}, where $I(Y_{true}; Y_{pred})$ denotes the mutual information between the true labels and predicted labels. $H(Y_{true})$ denotes the entropy of the true labels and $H(Y_{pred})$ denotes the entropy of the predicted labels. Higher NMI score indicates better clustering quality. The ground truth label for NMI computation is obtained from the cell type information provided by the authors who published the datasets. Since, the reported cell types are outcome of practical biological experiments, they can be considered as the noise free true labels. We have performed experiments with embedding layer dimensionality set at $2$, $10$, and $20$. As we can see from Figure \ref{fig:nmi_comparison}, the proposed method, scRAE, outperforms the current state-of-the-art methods irrespective of embedding dimension as measured by NMI score. The performance boost might be ascribed to the fact that scRAE is able to automatically operate at the optimal point on the bias-variance curve whereas the bias is high in methods such as scVI \cite{Lopez2018SCVI}, scVAE \cite{scVAE} and DR-A \cite{Lin2020}. scDeepCluster \cite{Tian2019scDeepCluster} and scziDesk \cite{scziDesk} adapt the methodologies laid in deep count autoencoder (DCA) \cite{Eraslan2019DCA} and deep embedding for clustering (DEC) \cite{dec}. These methods although avoid bias imposition through prior distribution, might over-fit the training data due to increased variance. Furthermore, the clustering layer in these two algorithms require the number of clusters as an input, which is not known \textit{a priori} in an unsupervised setting. Table \ref{tab:average_nmi} provides the average performance measured using NMI over all datasets to summarise the observations in the bar plot of Figure \ref{fig:nmi_comparison}. It is seen that on an average scRAE performs better than the baselines with respect to Normalized Mutual Information (NMI) score.\par
\begin{figure*}[!ht]
    \centering
    \includegraphics[trim=2 13 2 2, clip, keepaspectratio, width=\textwidth]{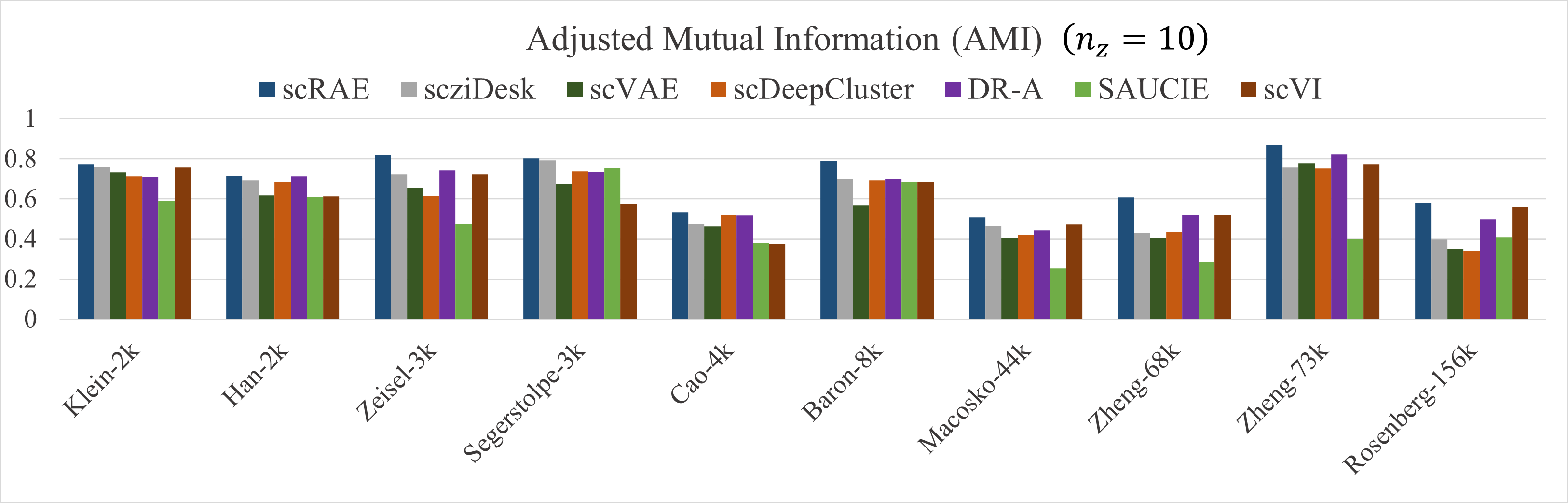}
    \caption{Clustering performance of scRAE is compared against $6$ deep learning based baseline methods. As can be seen from the figure, scRAE achieves highest AMI score for all of the ten datasets. Higher AMI score indicates better cluster purity and better performance w.r.t. ground truth labels.}
    \label{fig:ami_comparison}
    \vspace{-4mm}
\end{figure*}

Normalized Mutual Information, however, is not adjusted for chance. Generally, for two clustering assignments with a larger number of clusters, mutual information is higher, even when there is actually less shared information. Adjusted Mutual Information (AMI) takes the above fact into consideration and adjusts the Mutual Information (MI) score to account for chance. AMI is defined as follows:
\begin{equation}
    \text{AMI} = \frac{I(Y_{true}, Y_{pred}) - \mathbb{E}I(Y_{true}, Y_{pred})}{\frac{1}{2}(H(Y_{true}) + H(Y_{pred})) - \mathbb{E}I(Y_{true}, Y_{pred})}
\end{equation}
Figure \ref{fig:ami_comparison} compares AMI scores achieved by different deep learning based method when the embedding layer dimensionality is $10$. As before, scRAE outperforms the current state-of-the-art methods.\par

% Figure \ref{fig:preprocessing} presents the flowchart view of the preprocessing steps elaborated above. 
We have used APIs provided by the Scanpy \cite{Wolf2018} toolkit\footnote{\url{https://github.com/theislab/scanpy}} for data preprocessing.\par
Next we present the performance of the proposed scRAE for the two downstream tasks of clustering and visualization. We compare its performance against several state-of-the-art baseline methods in section \ref{sec:clustering_expt} and \ref{sec:data_vis_expt}.

\begin{figure*}[!ht]
    \centering
    \includegraphics[trim=2 6 2 2, clip, keepaspectratio, width=\textwidth]{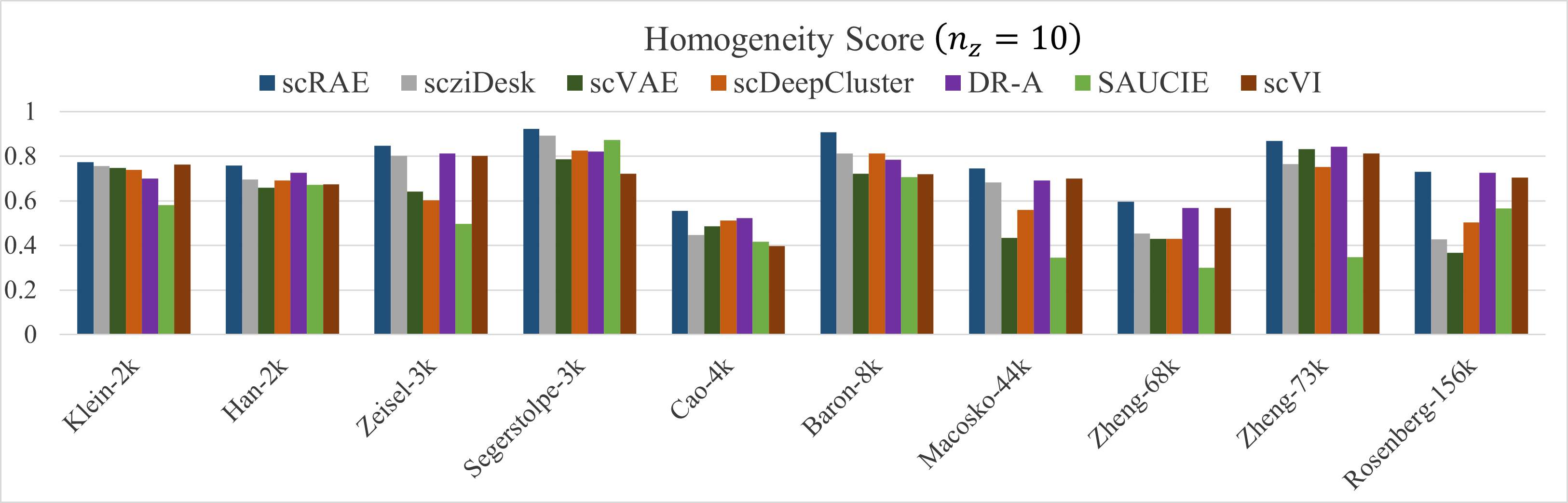}
    \caption{Clustering performance of scRAE is compared against $6$ deep learning based baseline methods. As can be seen from the figure, scRAE achieves highest Homogeneity Score (HS) for all of the ten datasets. Higher score is better.}
    \label{fig:hs_comparison}
    \vspace{-4mm}
\end{figure*}

\begin{figure*}[!ht]
    \centering
    \includegraphics[trim=2 5 2 2, clip, keepaspectratio, width=\textwidth]{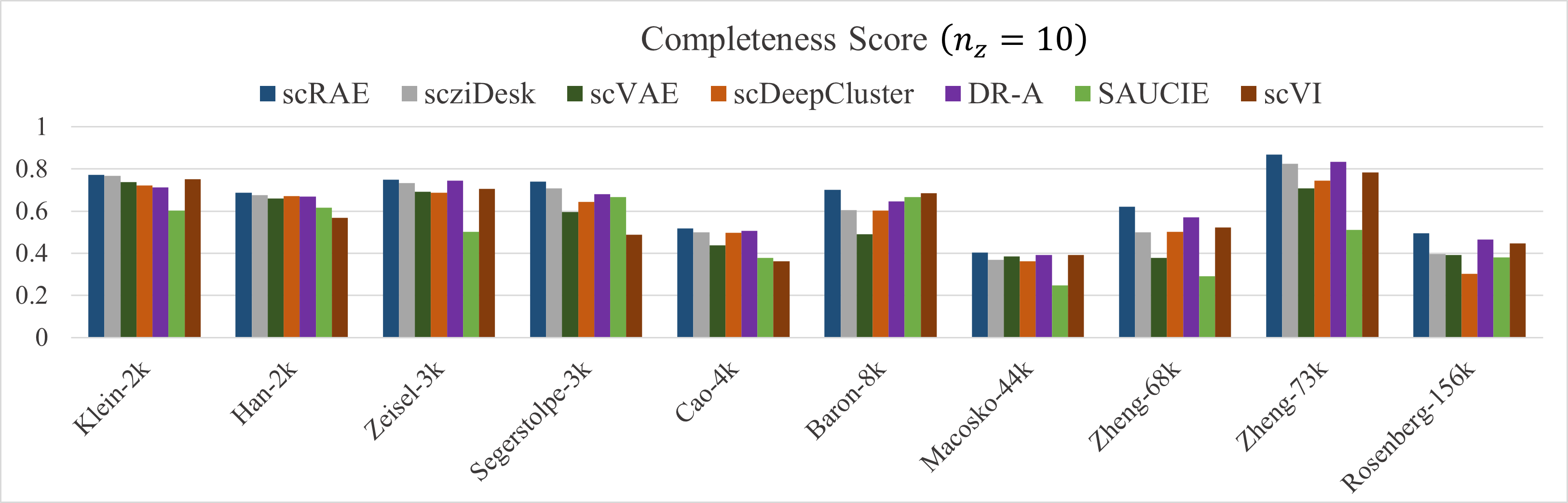}
    \caption{Clustering performance of scRAE is compared against $6$ deep learning based baseline methods. As can be seen from the figure, scRAE achieves highest Completeness Score (CS) for all of the ten datasets. Higher score is better.}
    \label{fig:cs_comparison}
    \vspace{-2mm}
\end{figure*}

\begin{figure*}[t!]
\vspace{-2mm}
    \centering
    \begin{subfigure}[t]{0.24\textwidth}
        \centering
        \includegraphics[trim={10 15 15 13}, clip, keepaspectratio, width=\textwidth]{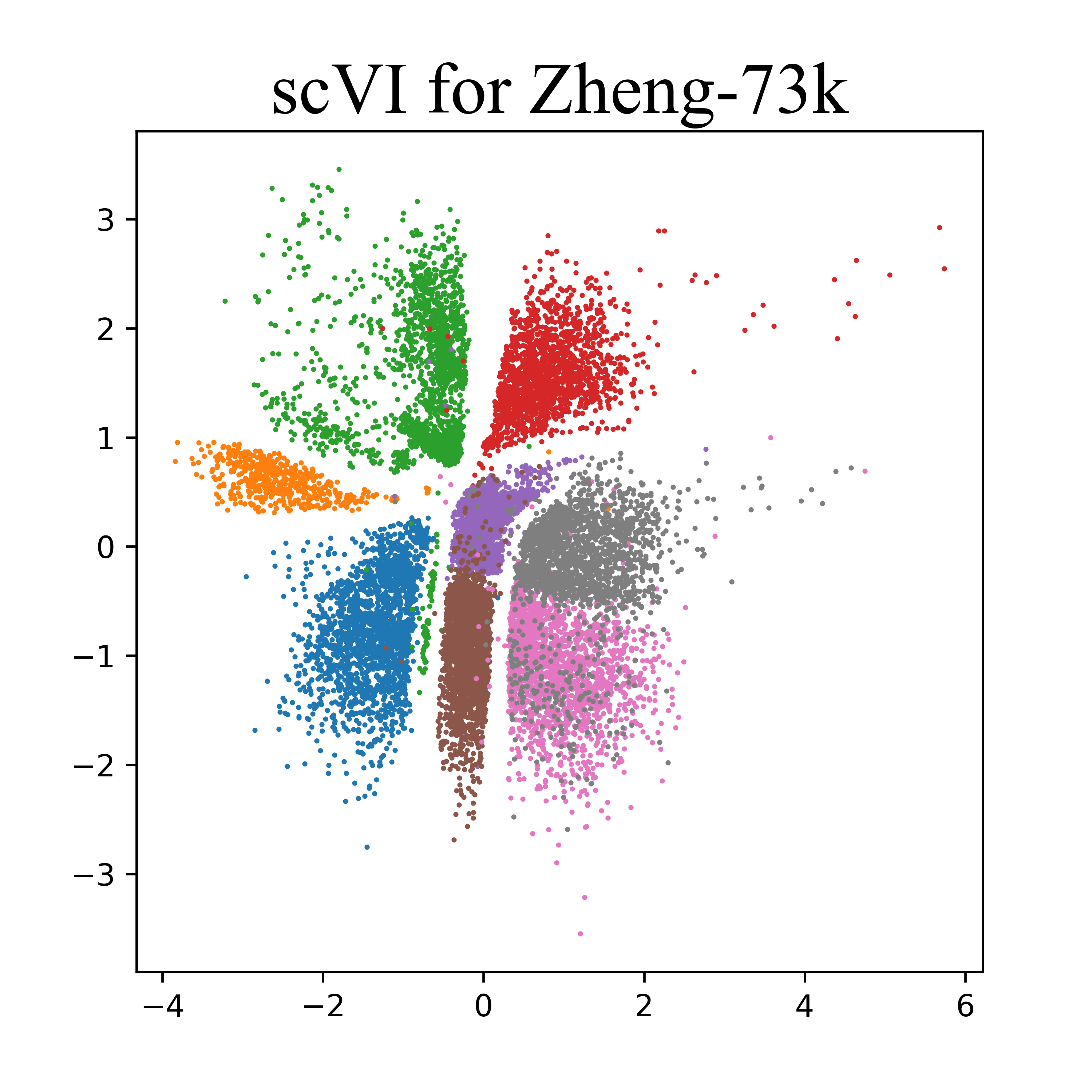}
        \caption{}
        \label{fig:scVI}
    \end{subfigure}%
    ~ 
    \begin{subfigure}[t]{0.24\textwidth}
        \centering
        \includegraphics[trim={10 15 15 13}, clip, keepaspectratio, width=\textwidth]{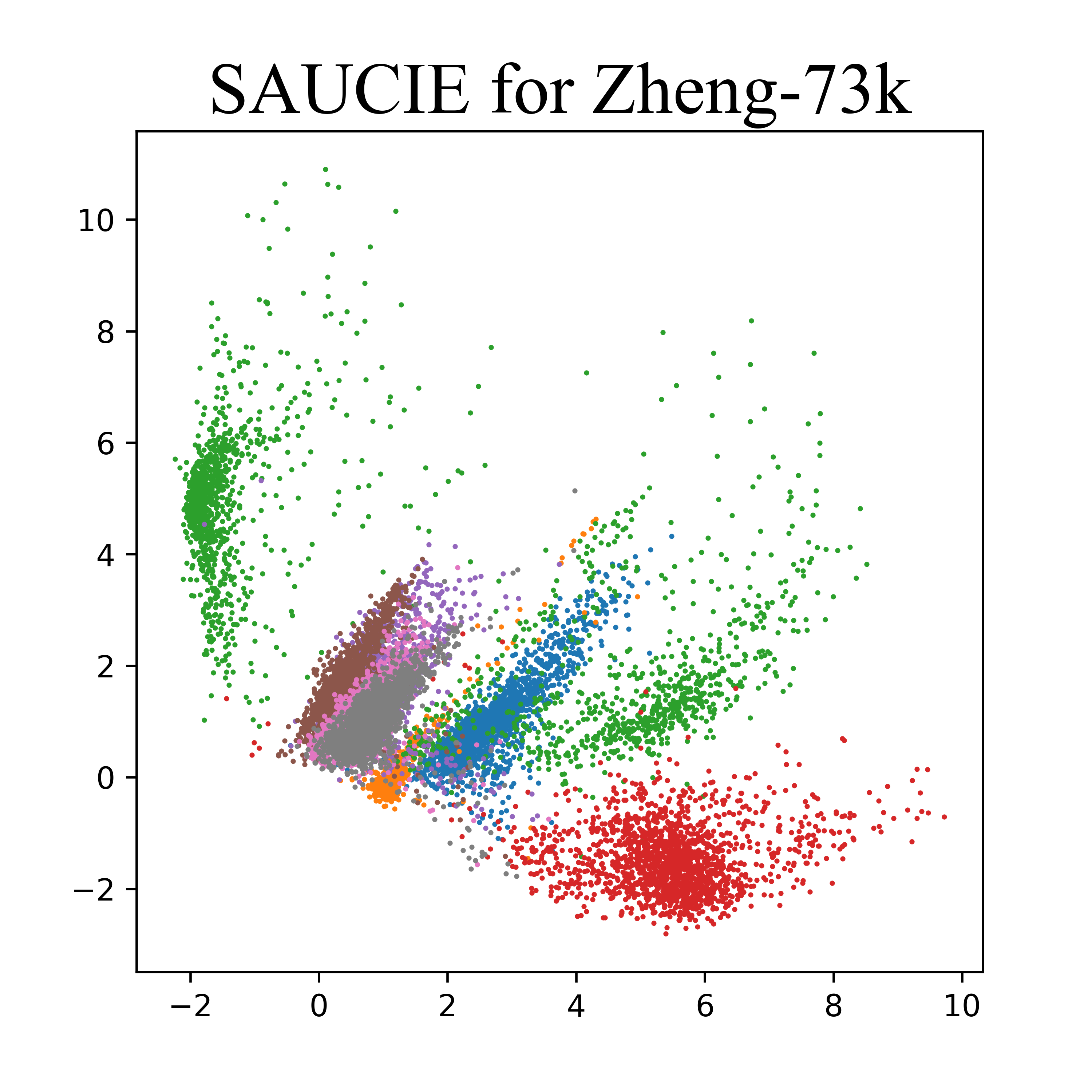}
        \caption{}
        \label{fig:SAUCIE}
    \end{subfigure}%
    ~
    \begin{subfigure}[t]{0.24\textwidth}
        \centering
        \includegraphics[trim={10 15 15 13}, clip, keepaspectratio,width=\textwidth]{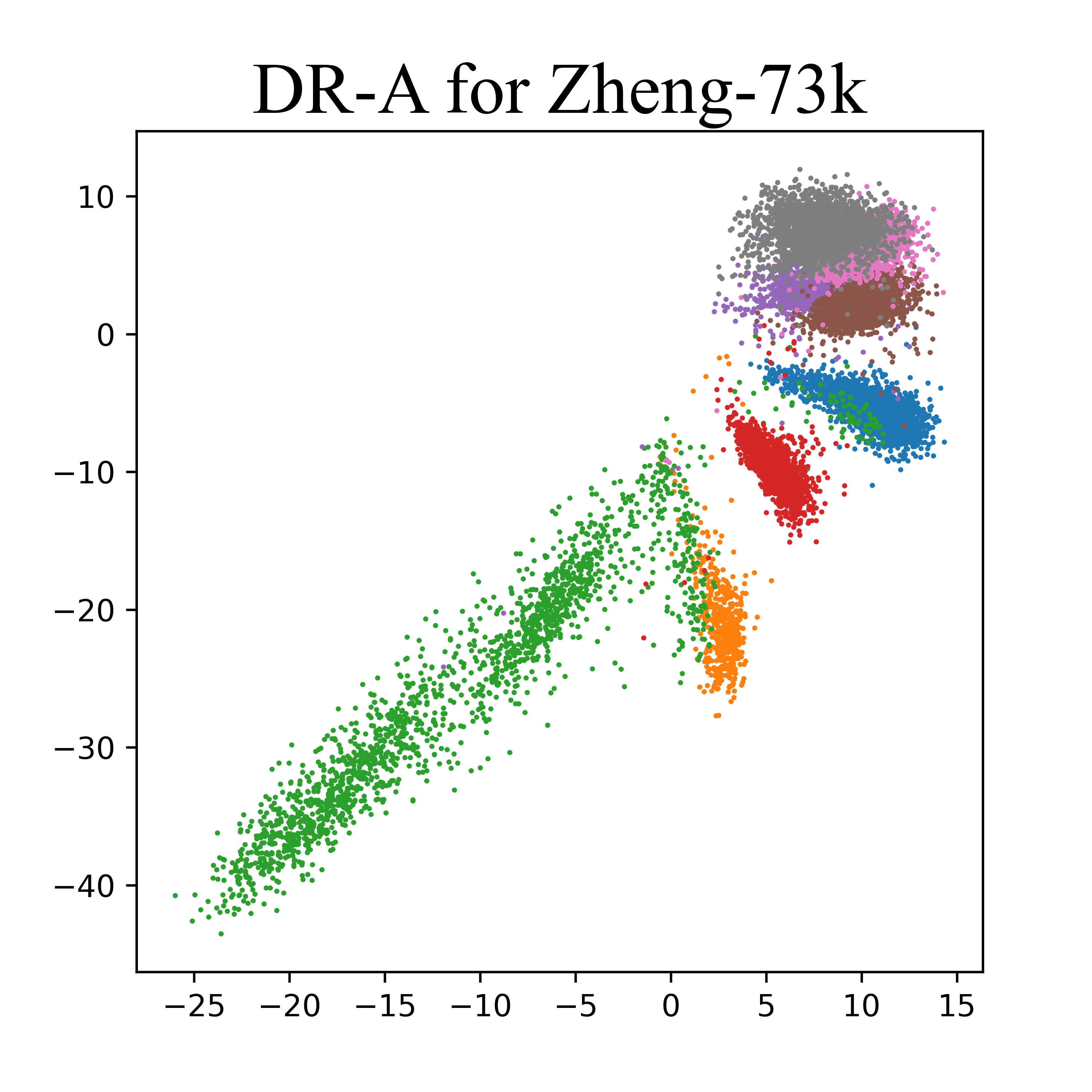}
        \caption{}
        \label{fig:DRA}
    \end{subfigure}%
    ~ 
    \begin{subfigure}[t]{0.24\textwidth}
        \centering
        \includegraphics[trim={10 15 15 13}, clip, keepaspectratio,width=\textwidth]{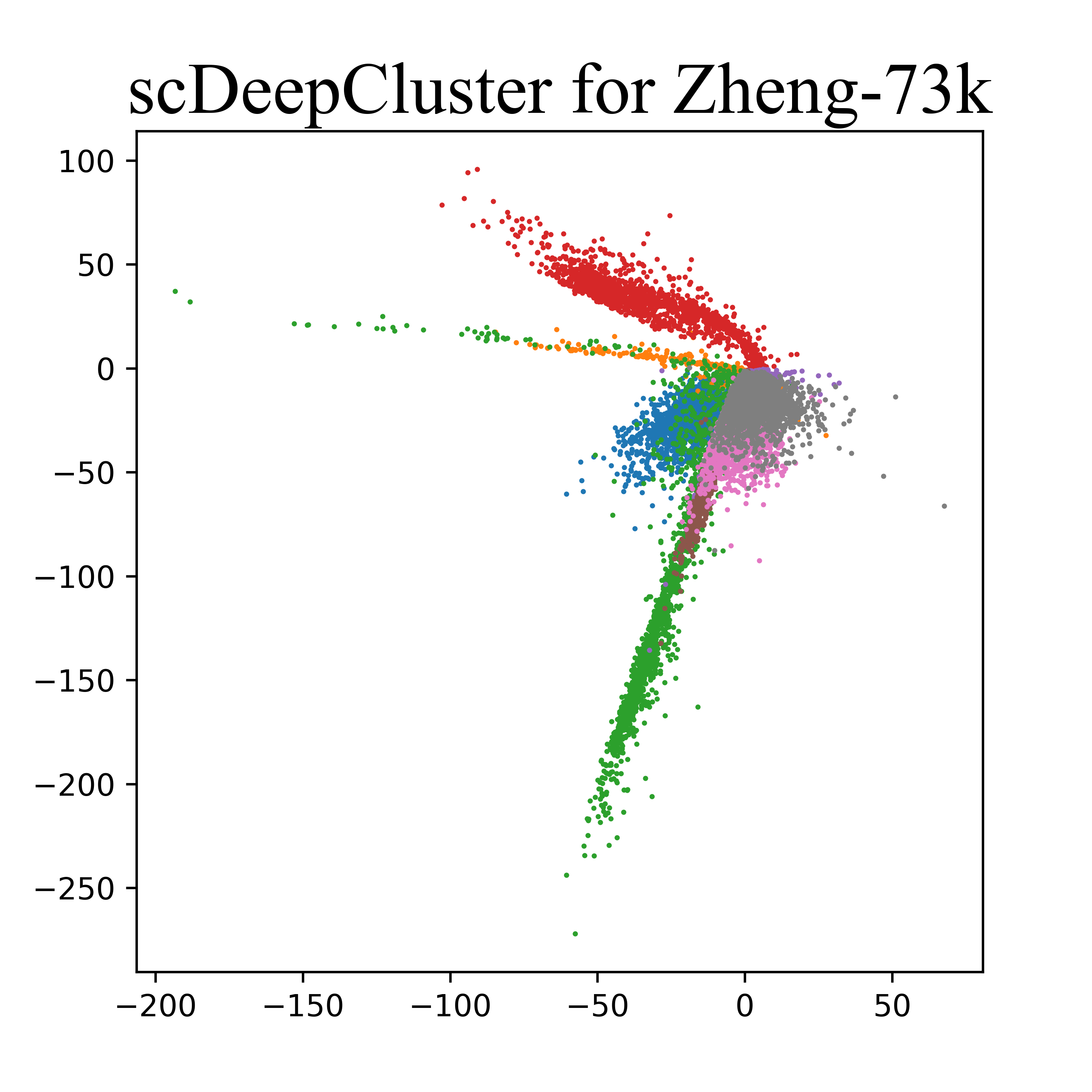}
        \caption{}
        \label{fig:scDeepCluster}
    \end{subfigure}%
    \hfill
    \begin{subfigure}[t]{0.24\textwidth}
        \centering
        \includegraphics[trim={10 15 15 13}, clip, keepaspectratio,width=\textwidth]{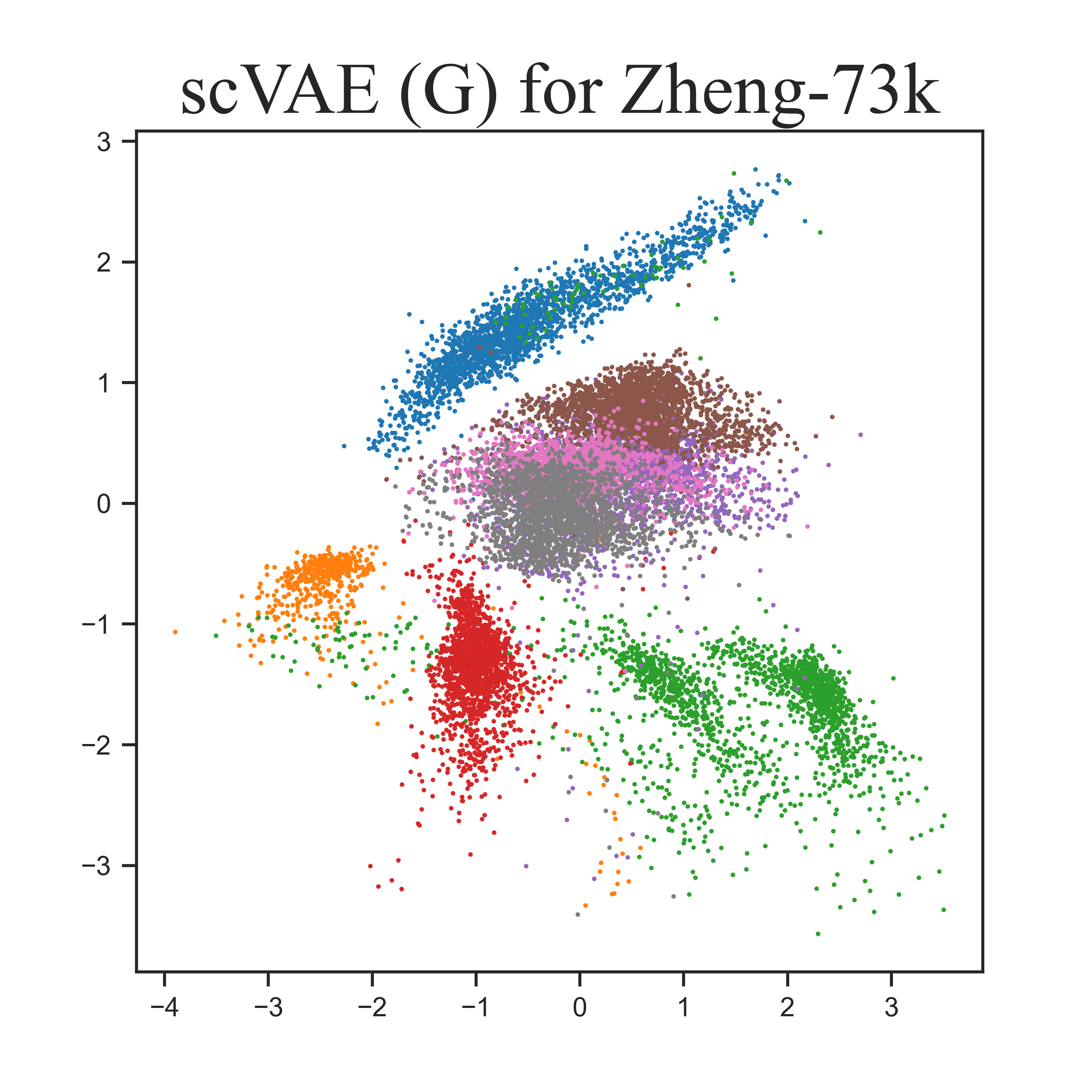}
        \caption{}
        \label{fig:scVAE_Gauss}
    \end{subfigure}%
    ~
    \begin{subfigure}[t]{0.24\textwidth}
        \centering
        \includegraphics[trim={10 15 15 13}, clip, keepaspectratio,width=\textwidth]{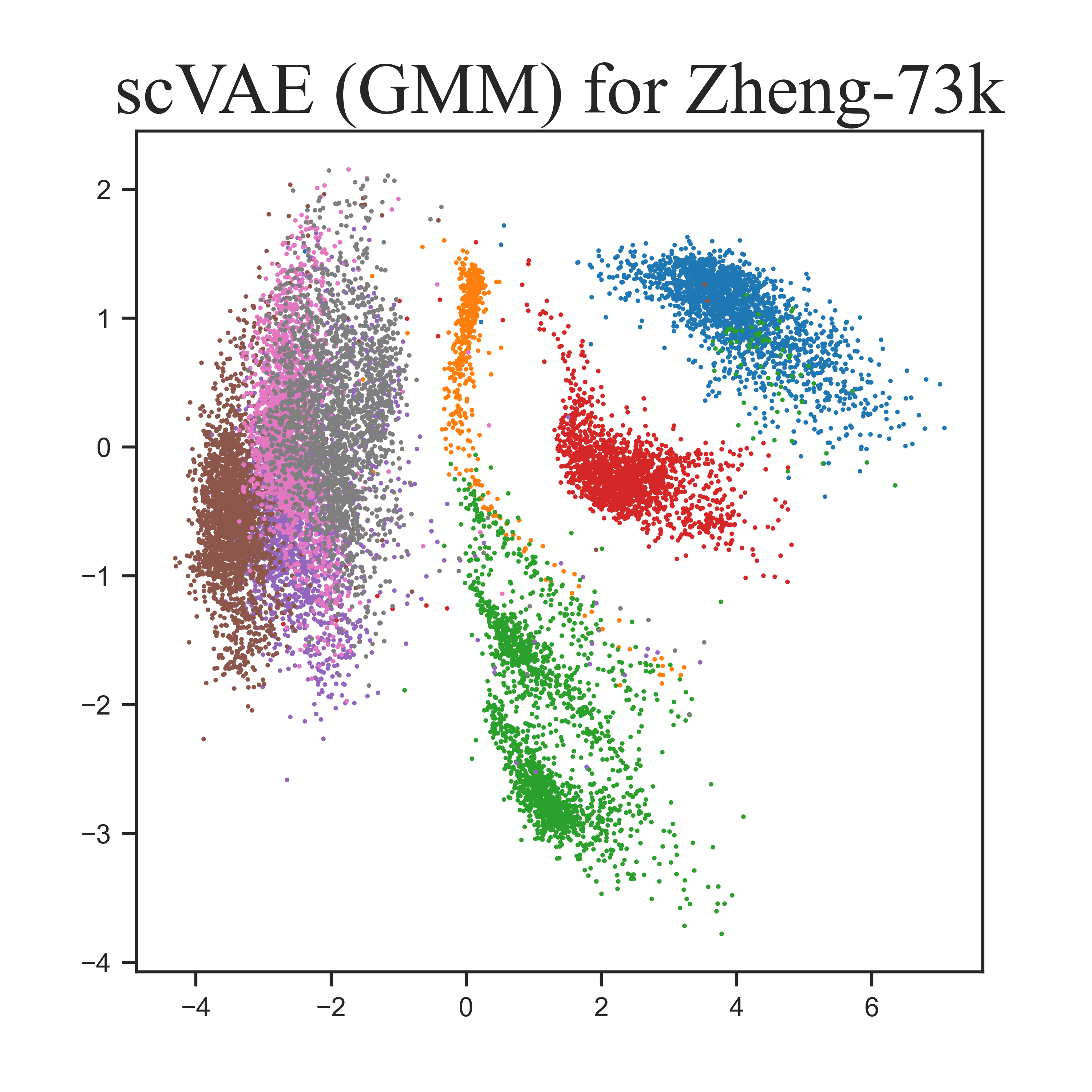}
        \caption{}
        \label{fig:scVAE_GMM}
    \end{subfigure}%
    ~
    \begin{subfigure}[t]{0.24\textwidth}
        \centering
        \includegraphics[trim={10 15 15 13}, clip, keepaspectratio,width=\textwidth]{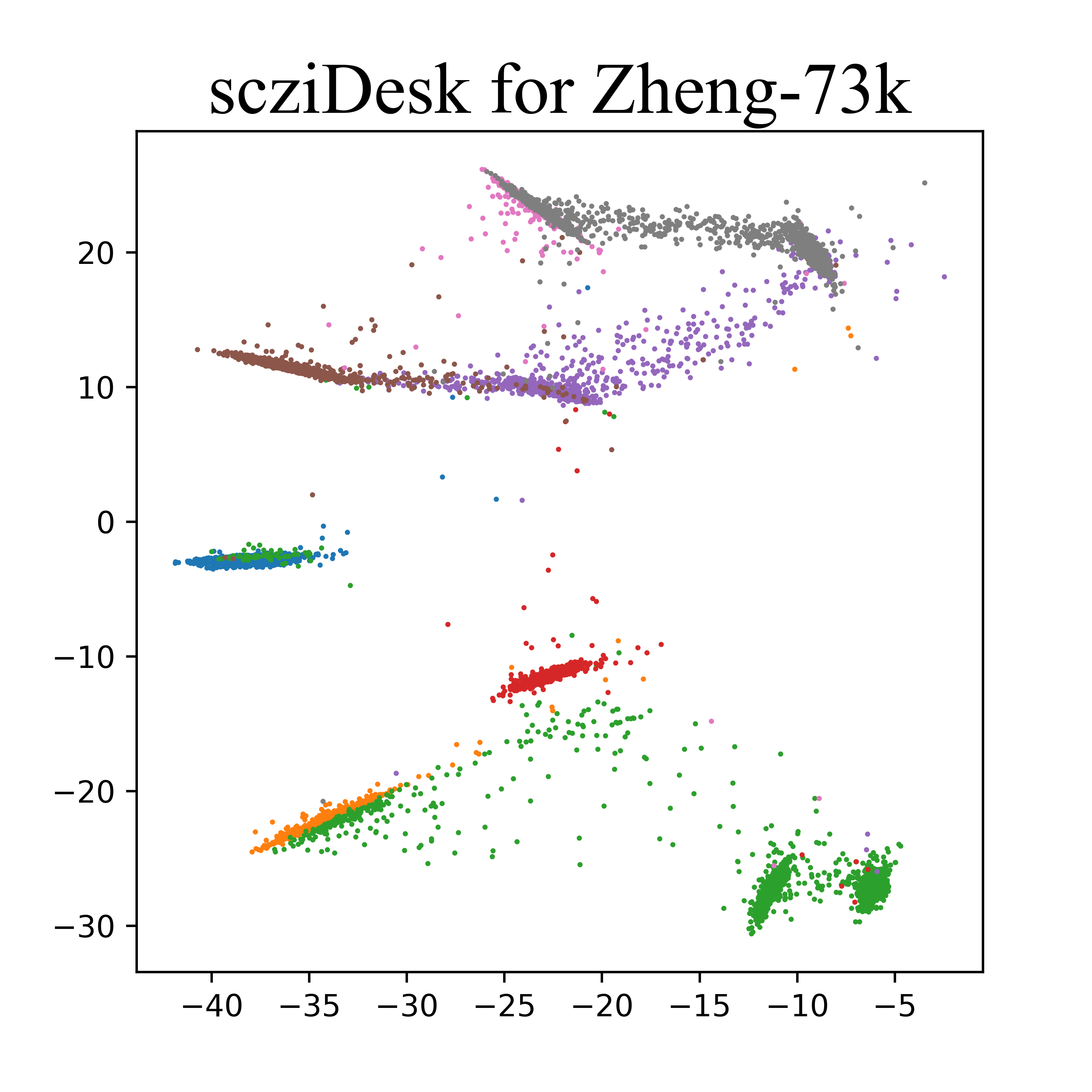}
        \caption{}
        \label{fig:scziDesk}
    \end{subfigure}%
    ~
    \begin{subfigure}[t]{0.24\textwidth}
        \centering
        \includegraphics[trim={10 15 15 13}, clip, keepaspectratio, width=\textwidth]{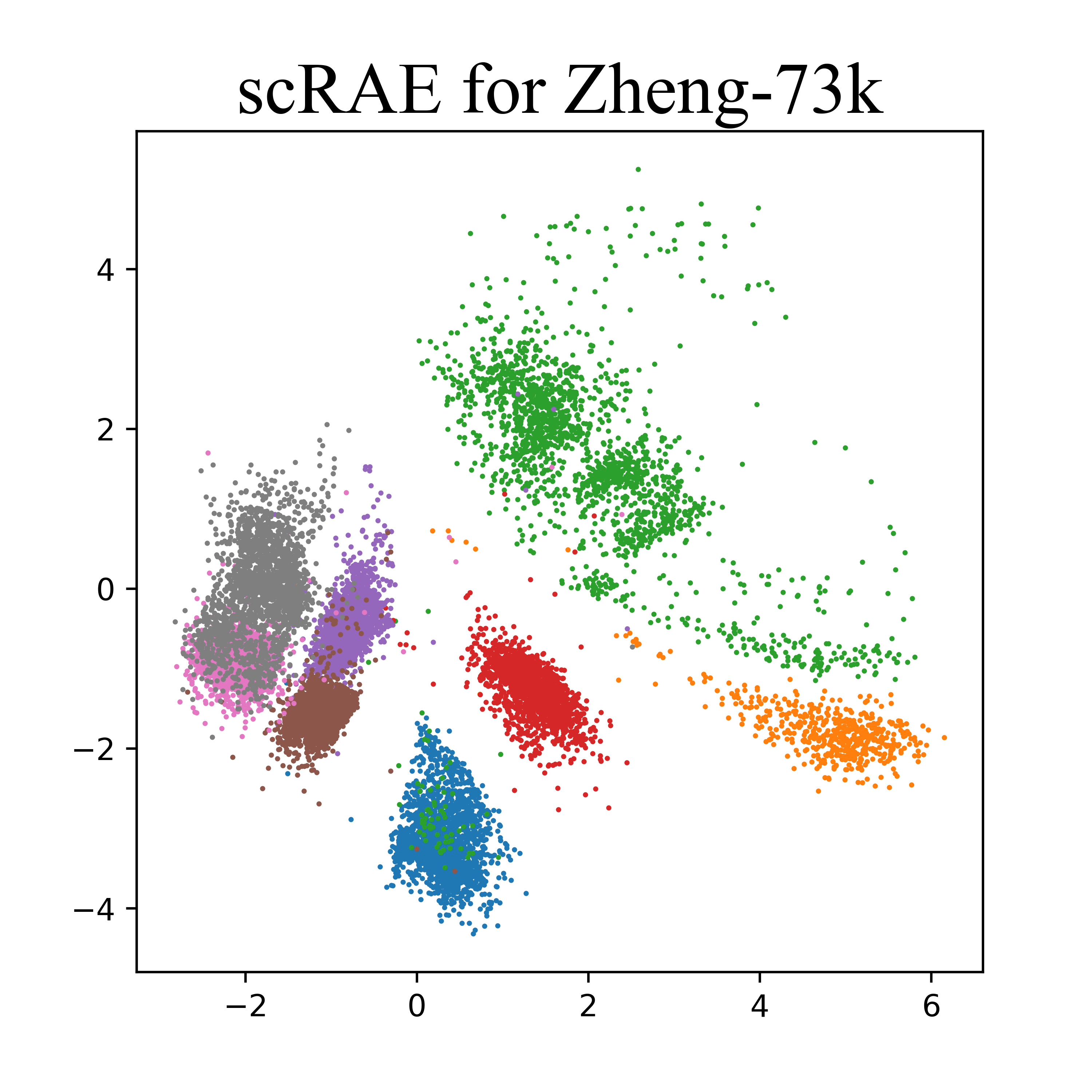}
        \caption{}
        \label{fig:scRAE}
    \end{subfigure}%
    \caption{Two dimensional Visualization for the dataset Zheng-$73k$ \cite{Zheng2017} using deep models such as (a) scVI \cite{Lopez2018SCVI}; (b) SAUCIE \cite{amodio2019exploring}; (c) DR-A \cite{Lin2020}; (d) scDeepCluster \cite{Tian2019scDeepCluster}; (e) scVAE \cite{scVAE} with Gaussian prior; (f)  scVAE \cite{scVAE} with mixture of Gaussian prior; (g) scziDesk \cite{scziDesk} and (h) scRAE (proposed). }
    \label{fig:zheng73k_visualization}
    \vspace{-2mm}
\end{figure*}
\begin{figure*}[ht!]
    \centering
    \begin{subfigure}[t]{0.33\textwidth}
        \centering
        \includegraphics[trim={10 15 15 13}, clip, keepaspectratio, width=\textwidth]{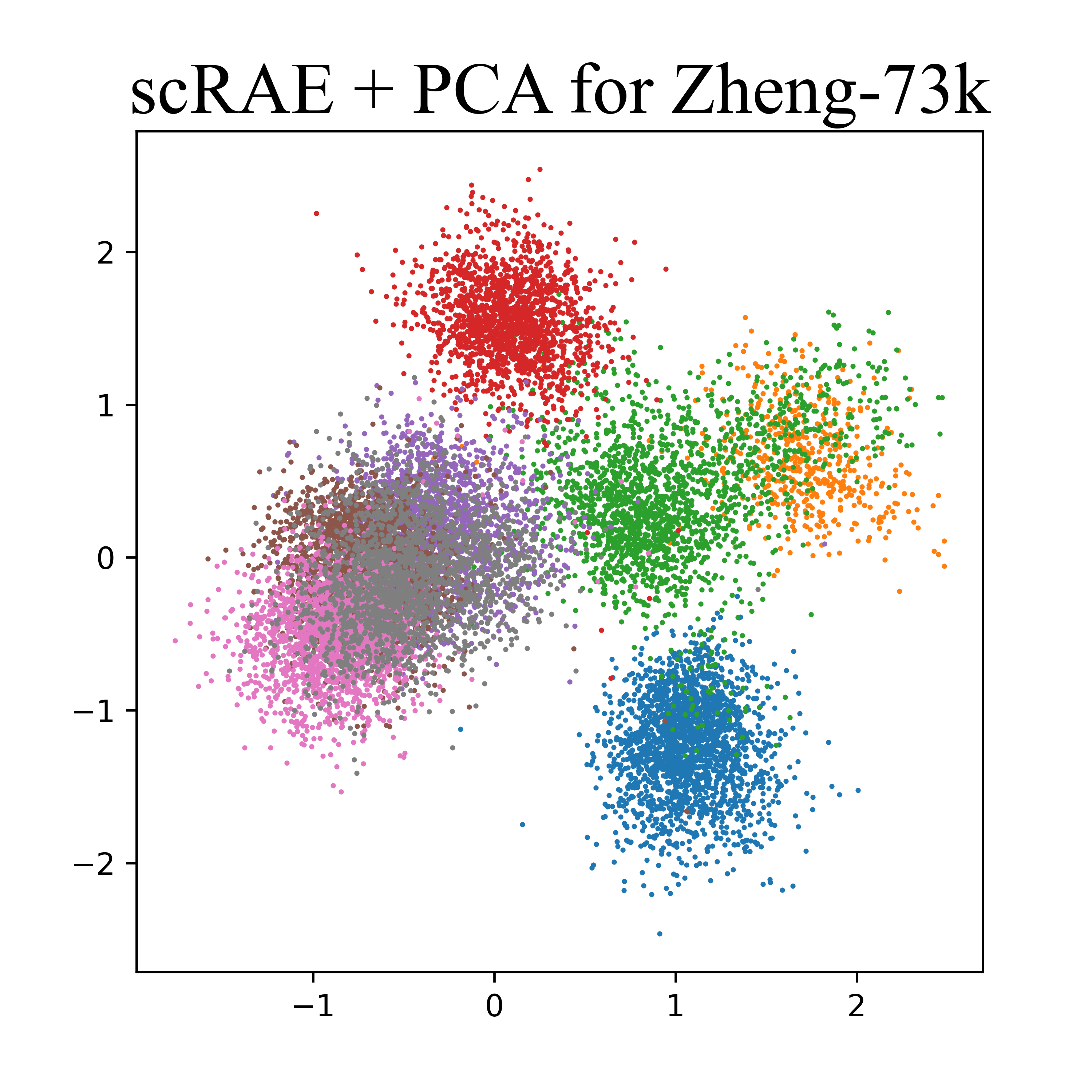}
        \caption{}
        \label{fig:scRAE_PCA}
    \end{subfigure}%
    ~ 
    \begin{subfigure}[t]{0.33\textwidth}
        \centering
        \includegraphics[trim={10 15 15 13}, clip, keepaspectratio, width=\textwidth]{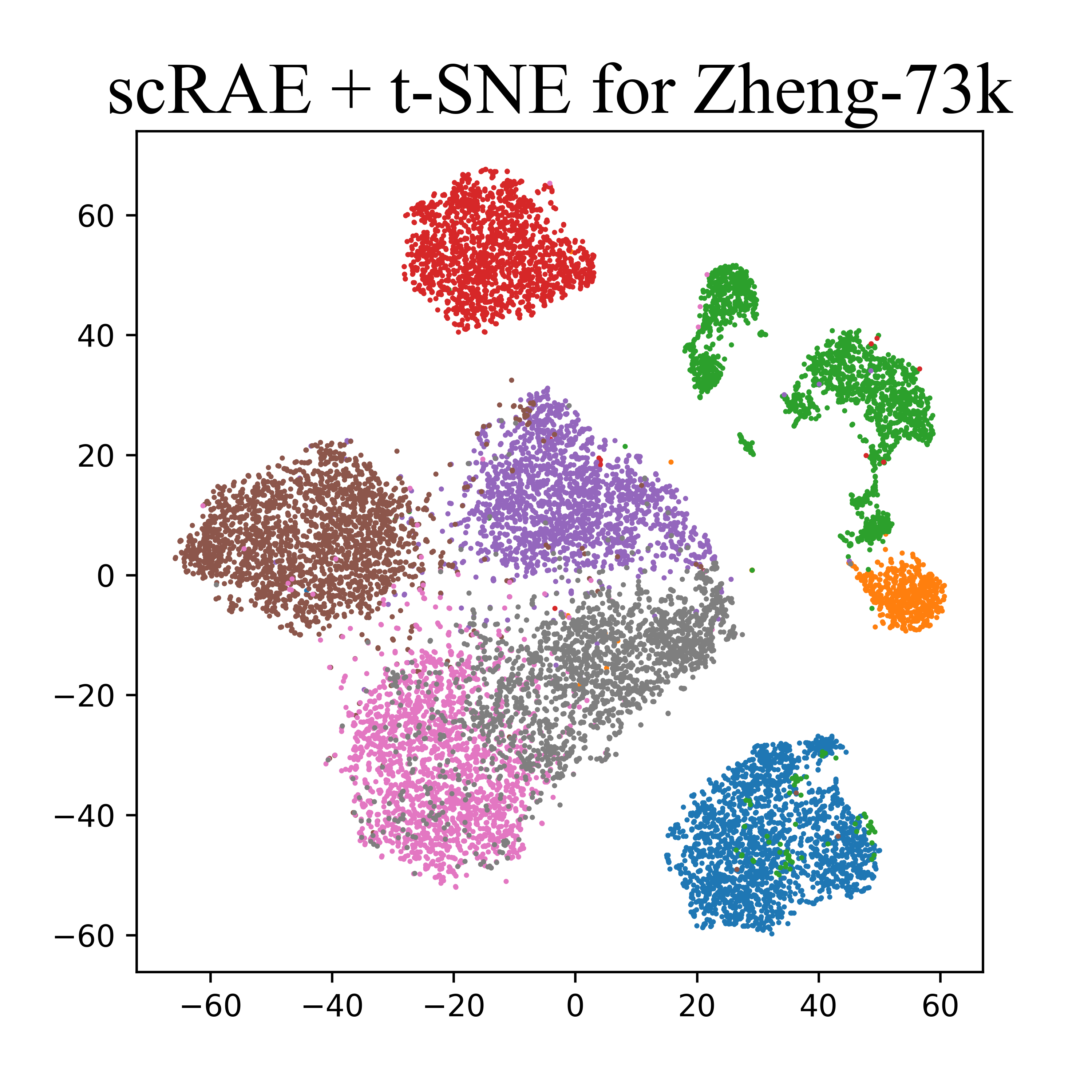}
        \caption{}
        \label{fig:scRAE_tSNE}
    \end{subfigure}%
    ~
    \begin{subfigure}[t]{0.33\textwidth}
        \centering
        \includegraphics[trim={10 15 15 13}, clip, keepaspectratio,width=\textwidth]{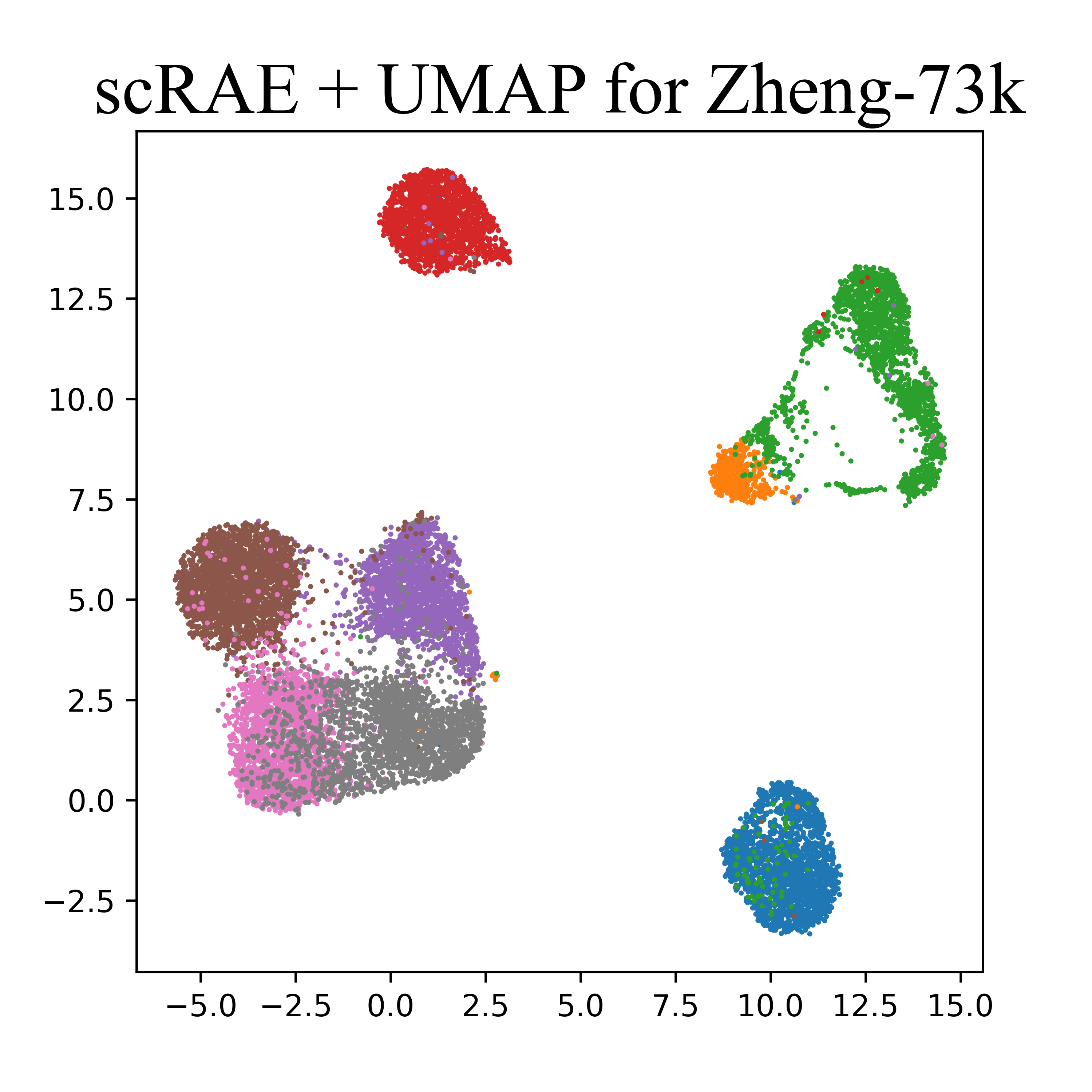}
        \caption{}
        \label{fig:scRAE_UMAP}
    \end{subfigure}%
    \hfill
    \caption{Visualization of the latent space of the proposed method, scRAE when the latent dimension is $10$ for the dataset Zheng-$73k$ \cite{Zheng2017} using (a) PCA \cite{Jolliffe2011}; (b) t-SNE \cite{tsne}; (c) UMAP \cite{mcinnes2018umap}.}
    \label{fig:Zheng73k_zDim10_visualization}
    \vspace{-2mm}
\end{figure*}

Figure \ref{fig:hs_comparison} and \ref{fig:cs_comparison} further qualitatively compares the proposed method against SOTA deep learning methods in terms of homogeneity score and completeness score. Higher homogeneity score achieved by scRAE indicates all of its clusters contain more data points which are members of a single class. Higher completeness score achieved by scRAE on the other hand indicates more data points that are members of a given class are elements of the same cluster.
\begin{figure*}[!ht]
    \centering
    \begin{subfigure}[t]{0.49\textwidth}
        \centering
        \includegraphics[trim={2 4 2 2}, clip, keepaspectratio, width=\textwidth]{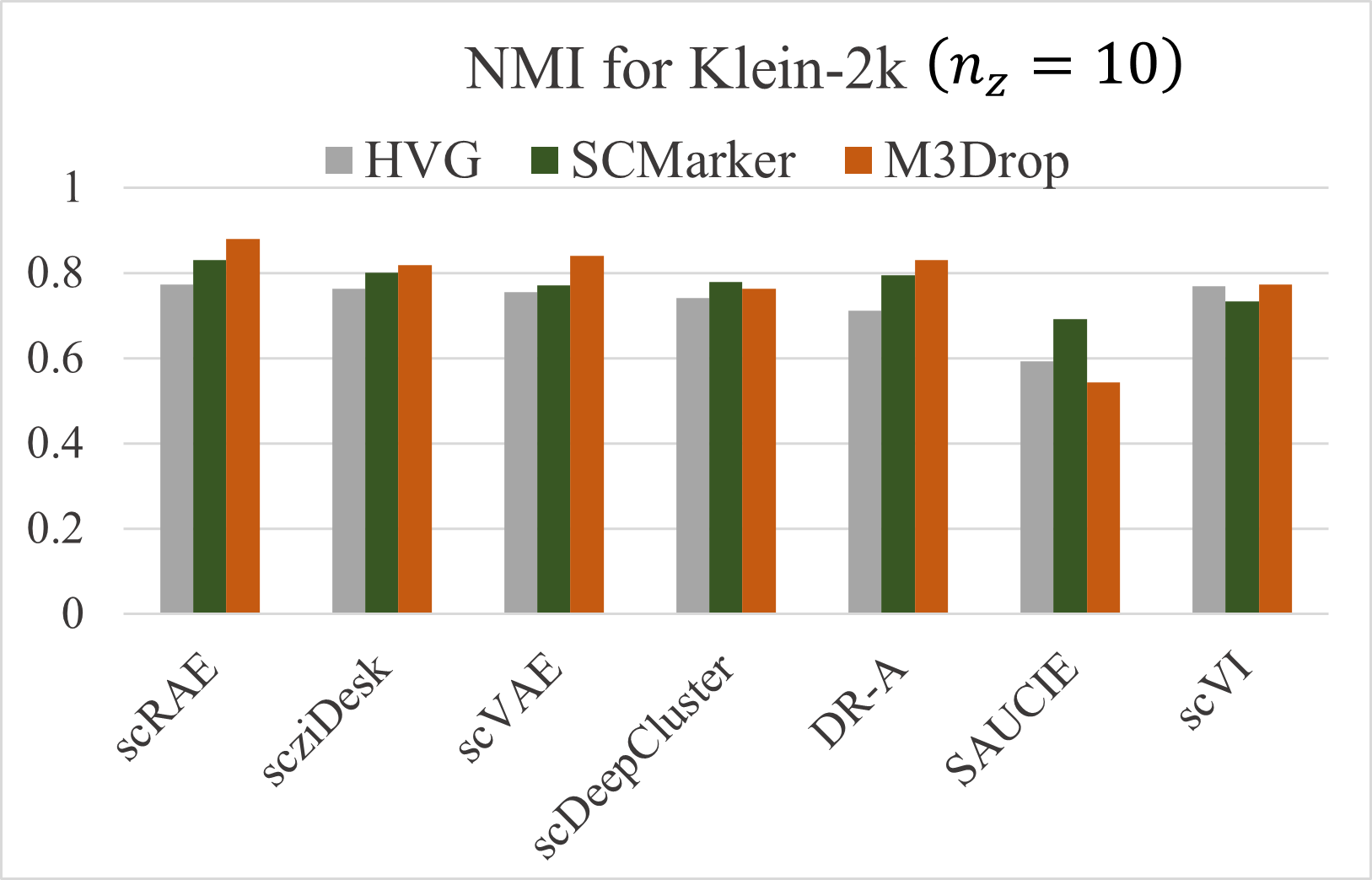}
        \caption{Impact of different gene selection method on clustering of Klein-$2k$ \cite{klein2015droplet}}
        \label{fig:klein_gene_selection}
    \end{subfigure}%
    ~~~
    \begin{subfigure}[t]{0.49\textwidth}
        \centering
        \includegraphics[trim={2 4 2 2}, clip, keepaspectratio, width=\textwidth]{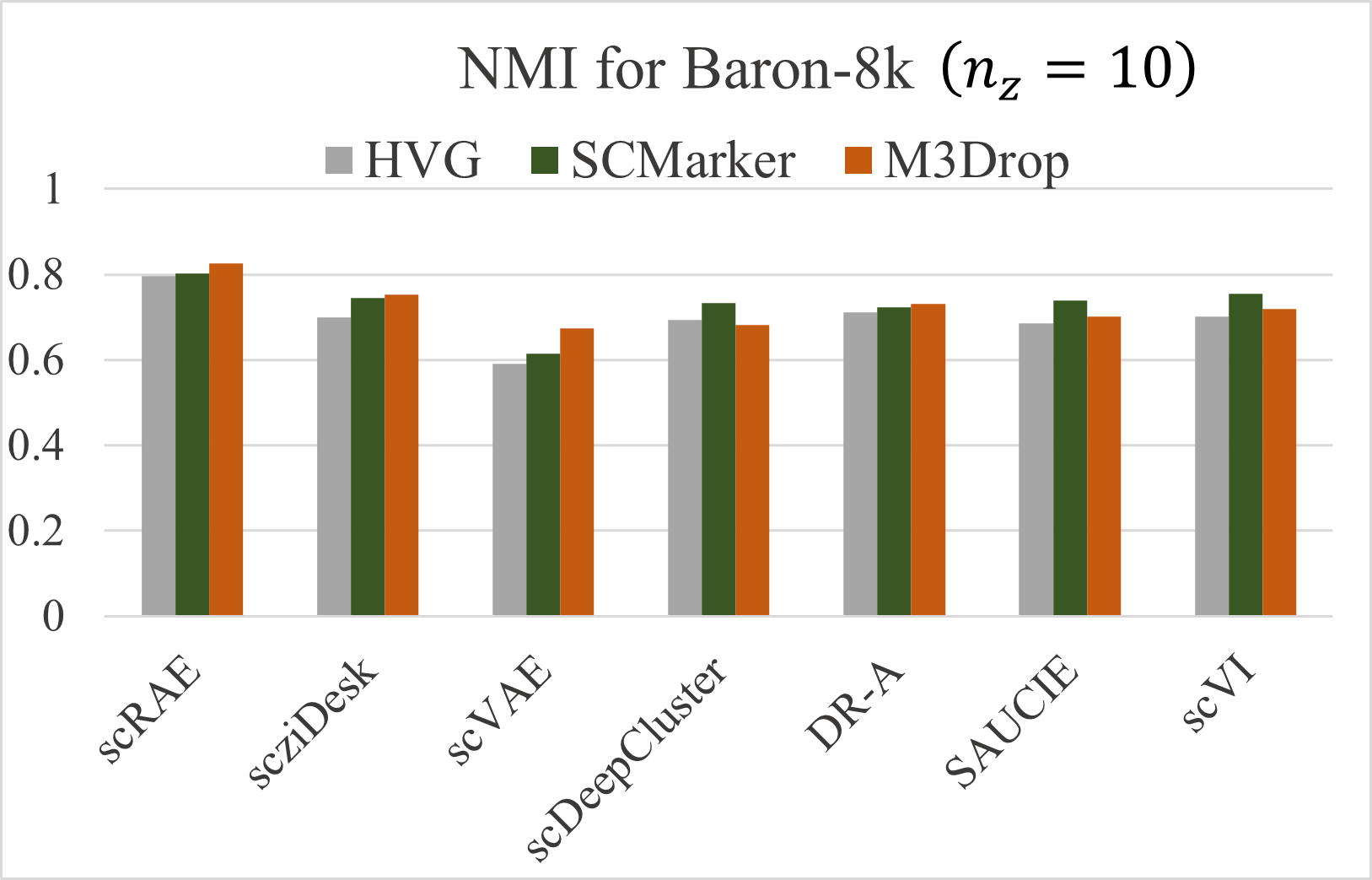}
        \caption{Impact of different gene selection method on clustering of Baron-$8k$ \cite{baron2016single}}
        \label{fig:baron_gene_selection}
    \end{subfigure}%
    \hfill
    \caption{Impact of gene selection method on clustering performance of different methods for two real-world datasets. Highest NMI score is achieved by scRAE when M3Drop \cite{M3Drop} is used for gene selection for both the datasets.}
    \label{fig:gene_selection}
\end{figure*}

\begin{table*}[!ht]
\begin{center}
\caption{Impact of gene selection method on clustering performance as measured by Normalized Mutual Information (NMI) score averaged over two real datasets Klein-$2k$ \cite{klein2015droplet} and Baron-$8k$ \cite{baron2016single}. In most of the cases, the clustering performance of an algorithm has improved when advance gene selection techniques such as SCMarker \cite{SCMarker} or M3Drop \cite{M3Drop} is used. Our proposed method scRAE outperforms current SOTA baselines for a fixed gene selection method and achieves best performance for M3Drop \cite{M3Drop}.}\label{tab:gene_selection_vs_nmi}
\begin{tabular}{cccccccc}
\hline
         & scRAE  & scziDesk \cite{scziDesk} & scVAE \cite{scVAE}  & scDeepCluster \cite{Tian2019scDeepCluster} & DR-A \cite{Lin2020}   & SAUCIE \cite{amodio2019exploring} & scVI \cite{Lopez2018SCVI}   \\
         \hline\hline
HVG      & \textbf{0.7849} & 0.7309   & 0.6731 & 0.7172        & 0.7111 & 0.6389 & 0.7349 \\
SCMarker & \textbf{0.8167} & 0.7719   & 0.6928 & 0.7556        & 0.7587 & 0.7159 & 0.7437 \\
M3Drop   & \textbf{0.8526} & 0.7859   & 0.7567 & 0.7215        & 0.7805 & 0.6219 & 0.7460 \\
\hline
\end{tabular}
\end{center}
\vspace{-3mm}
\end{table*}

\begin{table*}[!ht]
\begin{center}
\caption{Comparison of average run time (in seconds rounded off to closest integer) of different algorithms on the smallest and the largest out of ten real datasets considered in this work.}
\label{tab:runtime}
\begin{tabular}{crrrrrrr}
\hline
Dataset                                & scRAE & scziDesk \cite{scziDesk} & scVAE \cite{scVAE} & scDeepCluster   \cite{Tian2019scDeepCluster} & DR-A \cite{Lin2020} & SAUCIE \cite{amodio2019exploring} & scVI \cite{Lopez2018SCVI} \\
\hline\hline
Klein-$2k$   \cite{klein2015droplet}   & 273                       & 181                                          & 367                                    & 156                                                              & 279                                     & 150                                                     & 289                                           \\
Rosenberg-$156k$   \cite{Rosenberg176} & 1785                      & 2689                                         & 1956                                   & 2317                                                             & 1927                                    & 663                                                     & 2409           \\
\hline
\end{tabular}
\end{center}
\vspace{-6mm}
\end{table*}
\subsection{Data Visualization}\label{sec:data_vis_expt}
Finally, to evaluate the effectiveness of the proposed method qualitatively, we have performed some experiments to visualize the compressed representation. The purpose of these experiments is to visually identify different cell types. Figure \ref{fig:zheng73k_visualization} presents the two dimensional representations learnt by different deep learning based models for Zheng-$73k$ \cite{Zheng2017} dataset. Visualization is performed using the test split after training is complete. Once, the computation of two dimensional representations of the test examples are complete they are plotted directly in Figure \ref{fig:zheng73k_visualization}. The ground truth labels are used to color the similar cells with same color and different cell types with separate colors. As we can see in Figure \ref{fig:scRAE}, in the latent space of scRAE, the representations corresponding to similar cells are closely located while the representations corresponding to dissimilar cells are far apart. Besides, some clusters are split into several sub-clusters indicating either biological effect or batch effect.\par
In Figure \ref{fig:Zheng73k_zDim10_visualization}, we have adopted a two step procedure to visualize the learnt $10$-dimensional latent space in a scRAE model. As can be seen in Figure \ref{fig:scRAE_tSNE} and \ref{fig:scRAE_UMAP}, when t-SNE \cite{tsne} or UMAP \cite{mcinnes2018umap} is used in the second step for visualization, the clusters in the dataset becomes readily prominent. Similar to the $2$-dimensional case (Figure \ref{fig:scRAE}), similar cells are closer and dissimilar cells are further apart. However, PCA being a linear method fails to capture the cluster information as prominently as in Figure \ref{fig:scRAE_tSNE} and \ref{fig:scRAE_UMAP} in two dimensions (Figure \ref{fig:scRAE_PCA}).

\begin{table}[!ht]
\centering
\caption{Impact of feature dimension on clustering performance of scRAE as measured by NMI score for $n_z=2$.}
\label{tab:feature_dim_vs_nmi}
\resizebox{\columnwidth}{!}{
\begin{tabular}{ccccccc}
\hline
\multirow{2}{*}{Dataset} & \multicolumn{6}{c}{Feature Dimension}               \\ \cline{2-7} 
                         & 500    & 600    & 700    & 800    & 900    & 1000   \\ \hline
Klein-2k\cite{klein2015droplet}                 & 0.8052 & 0.8072 & 0.8281 & 0.8234 & 0.8390 & 0.8354 \\
Han-2k\cite{HAN2018MicorwellSeq}                   & 0.7346 & 0.7422 & 0.7409 & 0.7417 & 0.7288 & 0.7467 \\
Zeisel-3k\cite{Zeisel1138}                & 0.6814 & 0.6962 & 0.7338 & 0.7207 & 0.7186 & 0.7009 \\
Macosko-44k\cite{Macosko20151202}              & 0.4991 & 0.4849 & 0.5326 & 0.5507 & 0.4883 & 0.5158 \\ \hline
\end{tabular}
}
\vspace{-4mm}
\end{table}

\subsection{Impact of Feature Dimension on Clustering}\label{sec:feature_dim_vs_nmi}
In order to evaluate the effect of feature size on the clustering performance, we vary number of features between $500$ and $1000$ in step size of 100. As can be seen from Table \ref{tab:feature_dim_vs_nmi}, the clustering performance saturates beyond feature dimension $700$. Similar behaviour have been observed for $n_z=10$ and $20$. Further, we have obtained similar ordering in the performance metrics of the proposed method and the baseline models for a fixed number of features.

\subsection{Impact of Gene Selection Method} \label{gene_sel}
As discussed in Sec. \ref{sec:preprocessing}, selection of informative genes is a critical preprocessing step, which not only reduces computational complexity but could potentially boost clustering performance. In this section, we study the effect of advanced gene selection methods on clustering. We consider two gene selection strategies namely SCMarker \cite{SCMarker}, and M3Drop \cite{M3Drop}. SCMarker \cite{SCMarker} is an unsupervised ab initio marker selection method. It is based on two metrics 1) discriminative power of individual gene expressions and 2) mutually coexpressed gene pairs (MCGPs). M3Drop \cite{M3Drop} identify genes with unusually high numbers of zeros, also called ‘dropouts’, among their observations. As seen from Figure \ref{fig:klein_gene_selection} and \ref{fig:baron_gene_selection}, the clustering performance of scRAE improves significantly when SCMarker or M3Drop is used for gene selection as compared to Highly Variable Gene (HVG) selection method \cite{Satija2015SEURAT}. M3Drop provides the best performance. This might be ascribed to the advantage of using the dropout-rate over variance as the dropout-rate can be estimated more accurately due to much lower sampling noise. Similar performance boost is observed in other baseline methods as well when SCMarker or M3Drop is used for gene selection (refer to Figure \ref{fig:gene_selection} and Table \ref{tab:gene_selection_vs_nmi}).
\subsection{Complexity Analysis}
In this section, we compare the run time complexity of the proposed method, scRAE with the baseline methods used. We have used a machine with Intel\textsuperscript{\textregistered} Xeon\textsuperscript{\textregistered} Gold 6142 CPU, 376GiB RAM, and Zotac GeForce\textsuperscript{\textregistered} GTX 1080 Ti 11GB Graphic Card for all of our experiments. Table \ref{tab:runtime} reports the average runtimes of all the algorithms for experiments on Klein-$2k$ \cite{klein2015droplet}, Rosenberg-$156k$ \cite{Rosenberg176} (The largest and the smallest datasets, similar observations were observed on the others as well). Preprocessing is a one time process and common step for all the methods. Hence we do not include the time required for preprocessing in this analysis. As can be seen from Table \ref{tab:runtime}, for comparable model capacities, SAUCIE \cite{amodio2019exploring} is the fastest, however most of the time its performance is the worst (see Table \ref{tab:average_nmi}, \ref{tab:gene_selection_vs_nmi}, Figure \ref{fig:nmi_comparison}, \ref{fig:ami_comparison}, \ref{fig:hs_comparison}, \ref{fig:cs_comparison}). scDeepCluster \cite{Tian2019scDeepCluster} and scziDesk \cite{scziDesk} holds the second and the third position respectively in terms of computation speed for Klein-$2k$. However, for Rosenberg-$156k$, where the sample complexity is large, scRAE occupies second position. Hence, it can be concluded that scRAE provides considerable boost in the performance compared to baselines without taking significantly higher computational overheads. 
\section{Conclusion}
To conclude, in this work, we argue that there is a bias-variance trade-off with the imposition of any prior on the latent space in the finite data regime. We propose a model-based deep learning method scRAE, a generative AE for single-cell RNA sequencing data, which can potentially operate at different points of the bias-variance curve. Unlike previous deep learning-based generative modeling approaches, scRAE flexibly learns the prior while imposing restrictions on the latent space due to joint training of the P-GEN and the AE. This facilitates trading-off bias-variance on the fly and can potentially determine the optimum operating point on the bias-variance curve. We have empirically demonstrated scRAE's efficacy in clustering on ten real-world datasets and quantitatively compared its performance against several deep learning-based approaches. scRAE achieves the best performance as measured by metrics such as NMI, AMI, homogeneity score, and completeness score irrespective of the bottleneck layer's dimensionality. For bottleneck dimension, $n_z=10$, scRAE's average performance over ten datasets can be summarized as follows: average NMI is $0.7093$, average AMI is $0.6994$, average homogeneity score is $0.7706$, and average completeness score is $0.6555$. We have empirically established the effectiveness of scRAE for the visualization of high-dimensional scRNA-seq datasets. We have provided exhaustive ablation studies to examine the impact of feature dimension and the effect of gene selection methods. We have demonstrated scRAE's computational efficiency through extensive experimentation. Further, we have illustrated the scalability of the proposed method on several large datasets \cite{Rosenberg176, Zheng2017}. As more and more single-cell data becomes available, we expect more applications of our proposed method.
\section*{Acknowledgment}
The authors would like to thank Ajay Sailopal, IIT Delhi, for his help in running some of the baseline methods. Parag Singla is supported by the DARPA Explainable Artificial Intelligence (XAI) Program with number N66001-17-2-4032, the Visvesvaraya Young Faculty Fellowships by Govt. of India and IBM SUR awards. Himanshu Asnani acknowledges the support of Department of Atomic Energy, Government of India, under the project no. 12-R\&D-TFR-5.01-0500 and a gift from Adobe Research. Any opinions, findings, conclusions or recommendations expressed in this paper are those of the authors and do not necessarily reflect the views or official policies, either expressed or implied, of the funding agencies.
\bibliographystyle{ieeetr}
\bibliography{ref}

\begin{IEEEbiography}[{\includegraphics[width=1in,height=1.25in,clip,keepaspectratio]{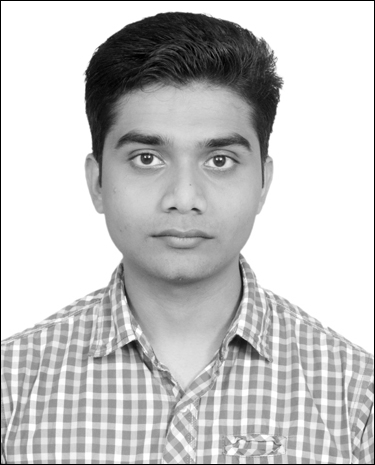}}]{Arnab Kumar Mondal} received his Bachelor of Engineering in Electronics and Telecommunication from Jadavpur University, India in 2013. Right after his graduation, he joined Centre for Development of Telematics (C-DOT), Delhi, and served as a research engineer there until July 2018. In C-DOT he had the opportunity to participate in cutting-edge projects such as the Dense Wavelength Division Multiplexing (DWDM) and Packet Optical Transport Platform (P-OTP). He joined IIT Delhi as a Ph.D. scholar in July, 2018 under the guidance of Prof. Prathosh AP and Prof. Parag Singla. His research interests lie primarily within the field of deep generative models and applied deep learning. He has published a couple of top-tier peer reviewed conference papers so far. His Ph.D. is supported by the Prime Minister's Research Fellows (PMRF) Scheme by Govt. of India.
\vspace{-10mm}
\end{IEEEbiography}
\begin{IEEEbiography}[{\includegraphics[width=1in,height=1.25in,clip,keepaspectratio]{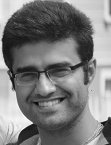}}]{Dr. Himanshu Asnani} is currently Reader (equivalent to tenure-track Assistant Professor) in the School of Technology and Computer Science (STCS) at the Tata Institute of Fundamental Research (TIFR), Mumbai and Affiliate Assistant Professor in the Electrical and Computer Engineering Department at University of Washington, Seattle. His research interests include information and coding theory, statistical learning and inference and machine learning. Dr. Asnani is the recipient of 2014 Marconi Society Paul Baran Young Scholar Award and was named Amazon Catalyst Fellow for the year 2018. He received his Ph.D. in Electrical Engineering Department in 2014 from Stanford University, working under Professor Tsachy Weissman, where he was a Stanford Graduate Fellow. Following his graduate studies, he worked in Ericsson Silicon Valley as a System Architect for couple of years, focusing on designing next generation networks with emphasis on network redundancy elimination and load balancing. Before joining TIFR, Dr. Asnani worked as a Research Associate in Electrical and Computer Engineering Department at University of Washington, Seattle. In the past, he has also held visiting faculty appointments in the Electrical Engineering Department at Stanford University and Electrical Engineering Department at IIT Bombay.
\vspace{-10mm}
\end{IEEEbiography}
\begin{IEEEbiography}[{\includegraphics[width=1in,height=1.25in,clip,keepaspectratio]{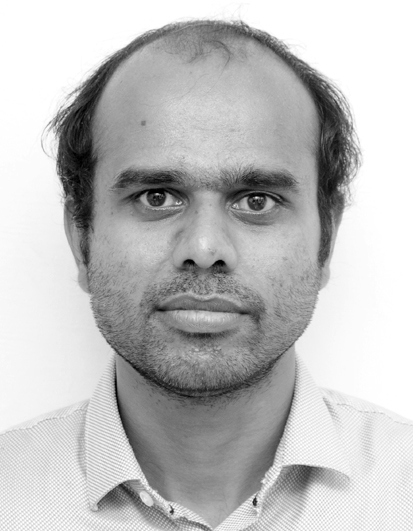}}]{Parag Singla} is an Associate Professor in the Department of
Computer Science and Engineering at IIT Delhi. He holds a Bachelors in Computer Science and Engineering (CSE) from IIT Bombay (2002) and Masters in CSE from University of Washington Seattle. He received his PhD from University of Washington Seattle in 2009 and did a PostDoc from University of Texas at Austin during the period 2010-11. He has been a faculty member in the Department of CSE at IIT Delhi since December 2011. Parag’s primary research interests lie in the areas of machine learning, specifically focusing on neuro symbolic reasoning. In the past, he has also worked extensively on graphical models and statistical relational AI. Parag has 40+ publications in top-tier peer reviewed conferences and journals. He also has one best paper award and two patents to his name. He is a recipient of the Visvesvaraya Young Faculty Research Fellowship by Govt. of India.
\vspace{-10mm}
\end{IEEEbiography}

\begin{IEEEbiography}[{\includegraphics[width=1in,height=1.25in,clip,keepaspectratio]{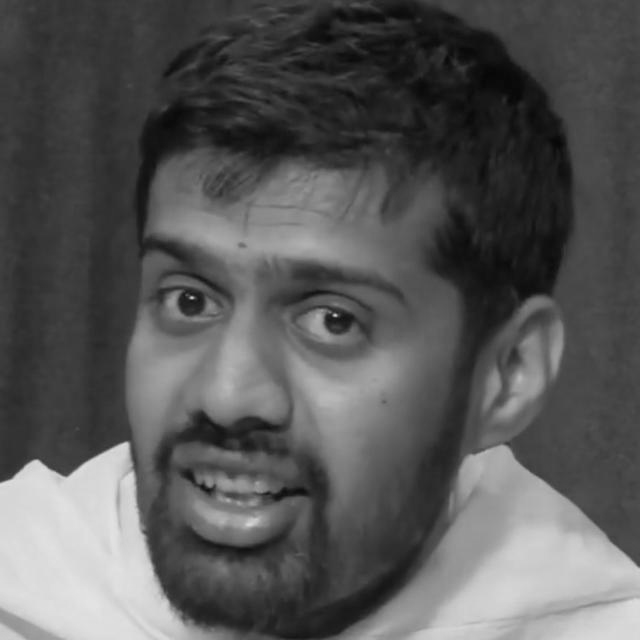}}]{Prathosh AP} is currently a faculty member at IIT Delhi, India. He received his Ph. D from Indian Institute of Science (IISc), Bangalore in 2015, in the area of temporal data analysis and applied machine learning. He submitted his PhD thesis in a record time of three years after his B.Tech. with many top-tier journal publications. Subsequently he worked in corporate research labs including Xerox Research India, Philips research and a start-up in CA, USA. His work in industry led to generation of several IP, comprising 10 granted US patents, most of which are commercialised. He joined IIT Delhi as an Assistant Professor in the computer technology group of Electrical Engineering where is currently engaged in research and teaching machine and deep learning courses. His current research includes informative-prior guide representational learning, deep generative models, cross-domain learning and their applications.
\end{IEEEbiography}

\end{document}

\title{scRAE: Deterministic Regularized Autoencoders with Flexible Priors for Clustering Single-cell Gene Expression Data\\\Large{Supplementary Material}}

\author{Arnab~Kumar~Mondal\textsuperscript{\textasteriskcentered}, Himanshu~Asnani\textsuperscript{\textdagger}, Parag~Singla\textsuperscript{\textasteriskcentered}, and Prathosh~AP\textsuperscript{\textasteriskcentered}
\thanks{\textasteriskcentered~indicated authors are affiliated with IIT Delhi. \textdagger~indicated author is affiliated with TIFR, Mumbai.}%
}

% The paper headers
\markboth{IEEE/ACM Transactions on Computational Biology and Bioinformatics}%
{Mondal \MakeLowercase{\textit{et al.}}: scRAE: Deterministic RAEs with Flexible Priors for Clustering Single-cell Gene Expression Data}

% make the title area
\maketitle

\section{Training Algorithm}

\begin{algorithm}[!h]
    \caption{Pseudo code for the training loop of scRAE \label{alg:scRAE-training-loop}}
    \hspace*{\algorithmicindent} \textbf{Hyper-parameters:}\\ Learning rates: $\eta_{AE}$, $\eta_{Critic}$, $\eta_{Gen}$, $\eta_{Enc}$.\\ Optimizers and its parameters:\\ $\text{AE\_OPT} = \text{Adam}(\text{lr}=\eta_{AE}, \beta_1=0.9, \beta_2=0.999)$, $\text{CRITIC\_OPT} = \text{Adam}(\text{lr}=\eta_{Critic}, \beta_1=0.0, \beta_2=0.9)$, $\text{GEN\_OPT} = \text{Adam}(\text{lr}=\eta_{Gen}, \beta_1=0.0, \beta_2=0.9)$, $\text{ENC\_OPT} = \text{Adam}(\text{lr}=\eta_{Enc}, \beta_1=0.0, \beta_2=0.9)$, $disc\_training\_ratio=5$. \\
    \begin{algorithmic}[1]
        \Function{Train}{}
            \For{$i \gets 1 \textrm{ to } training\_steps$}
                \State Minimize $L_{AE}$ and Update $\phi,~\theta$
                \If{$i \% disc\_training\_ratio$ == 0}
                    \State Minimize $L_{Gen}$ and Update $\psi$
                    \State Minimize $L_{Enc}$ and Update $\phi$
                \Else
                    \State Minimize $L_{Critic}$ and Update $\kappa$
                \EndIf
            \EndFor
        \EndFunction
    \end{algorithmic}
\end{algorithm}